\documentclass[lettersize,journal]{IEEEtran}
\usepackage{amsmath,amsfonts}
\usepackage{algorithmic}
\usepackage{algorithm}
\usepackage{array}
\usepackage[caption=false,font=footnotesize,labelfont=sf,textfont=sf]{subfig}
\usepackage{textcomp}
\usepackage{stfloats}
\usepackage{url}
\usepackage{verbatim}
\usepackage{graphicx}
\usepackage{cite}
\usepackage[normalem]{ulem}
\usepackage{hyperref}
\usepackage{booktabs}
\usepackage{tabularx}
\usepackage{multirow}
\usepackage{xcolor}
\begin{document}

\newcommand*\nnWNet{nn\sout{W}Net}
\newcommand*\nnMNet{nn\sout{M}Net}
\newcommand*\nnMNetPlus{\nnMNet{}\(^{\dagger}\)}

\title{\nnMNet{}: Baseline for Martian Terrain Semantic Segmentation}

\author{Ming-Han Lee, Chi-Yeh Chen
  \thanks{Ming-Han Lee is with the Department of Computer Science and Information
    Engineering, National Cheng Kung University, Tainan 701, Taiwan (e-mail:
    dereklee0310@gmail.com).}
  \thanks{Chi-Yeh Chen is with the Department of Computer Science and Information
    Engineering, National Cheng Kung University, Tainan 701, Taiwan (e-mail:
    chency@csie.ncku.edu.tw).}%
}



\maketitle

\begin{abstract}
  Semantic segmentation is a crucial task for understanding Mars, the most Earth-like planet in our solar system. However, it is challenging because the Martian surface is highly unstructured and complex, making accurate pixel-level prediction and fine-grained annotation difficult. Recent advancements in deep learning have introduced numerous methods and datasets to address these challenges. Nevertheless, the field lacks a robust, publicly available, and reproducible baseline, as well as a unified benchmark to facilitate fair evaluations. In this work, we present \nnMNet{}, a new baseline model designed for Martian terrain semantic segmentation. Building upon \nnWNet{}, we integrate linear attention to better capture global context and employ lightweight convolutions to reduce computational overhead. To bridge the gap between local and global representations, we introduce the Spatially-Aware Fusion Block (SAFB), which augments and combines features with diverse characteristics. Furthermore, we establish a new benchmark by curating and standardizing three high-quality datasets for thorough evaluation. \nnMNet{} achieves new state-of-the-art 86.61\%, 83.25\%, and 88.24\% mIoU on SynMars-TW, SynMars-Air, and MarsScapes, respectively. Our code, models, and datasets are publicly available at \url{https://github.com/dereklee0310/nnMNet}.
\end{abstract}


\section{Introduction}
\IEEEPARstart{M}{artian} terrain semantic segmentation is a fundamental task in Mars exploration.
Unlike image classification or object detection, semantic segmentation requires an in-depth understanding of the scene to make precise pixel-level predictions, which are essential for downstream applications such as geological analysis, path planning, and autonomous rover navigation~\cite{MarsReview}. However, Martian environments are far more challenging to segment than terrestrial scenes due to their highly unstructured, complex surfaces. Frequent local dust storms and global dust events (GDEs)~\cite{DustStorm} also lower the visibility of Mars rovers and make it difficult to acquire high-quality images. In short, accurately delineating topographical boundaries and identifying terrain categories with limited data remain significant challenges.

While deep learning is widely used for Martian terrain semantic segmentation, several issues hinder its practical applicability. First, state-of-the-art methods predominantly adopt Vision Transformer (ViT) architectures~\cite{ViT} for better global perception capabilities~\cite{RockFormer,MarsFormer,Light4Mars,AirFormer,LisseMars}. However, ViTs typically employ aggressive downsampling and upsampling (e.g., by a factor of four) to reduce computational cost~\cite{SwinTransformer,PVT,PVTv2,SegFormer}, which inevitably discards fine-grained local details. This limited local perception severely impairs the segmentation accuracy of small yet safety-critical objects, such as small rocks and gravel, posing potential risks to rover navigation. Second, existing methods are built on different frameworks with customized training configurations. The lack of standardization makes it difficult to discern whether performance gains stem from architectural designs or training strategies. Third, inconsistencies in annotation criteria and procedures across datasets introduce ambiguous labels, imprecise geological boundaries, and missing objects, degrading overall data quality. Many of these datasets also fail to provide clearly defined training, validation, and testing splits, making fair benchmarking difficult. Finally, the source code, pre-trained model weights, and datasets of most existing methods are proprietary, hindering reproducibility and restricting further development in both scientific research and real-world deployment.

\begin{figure}[t]
  \includegraphics[width=\linewidth]{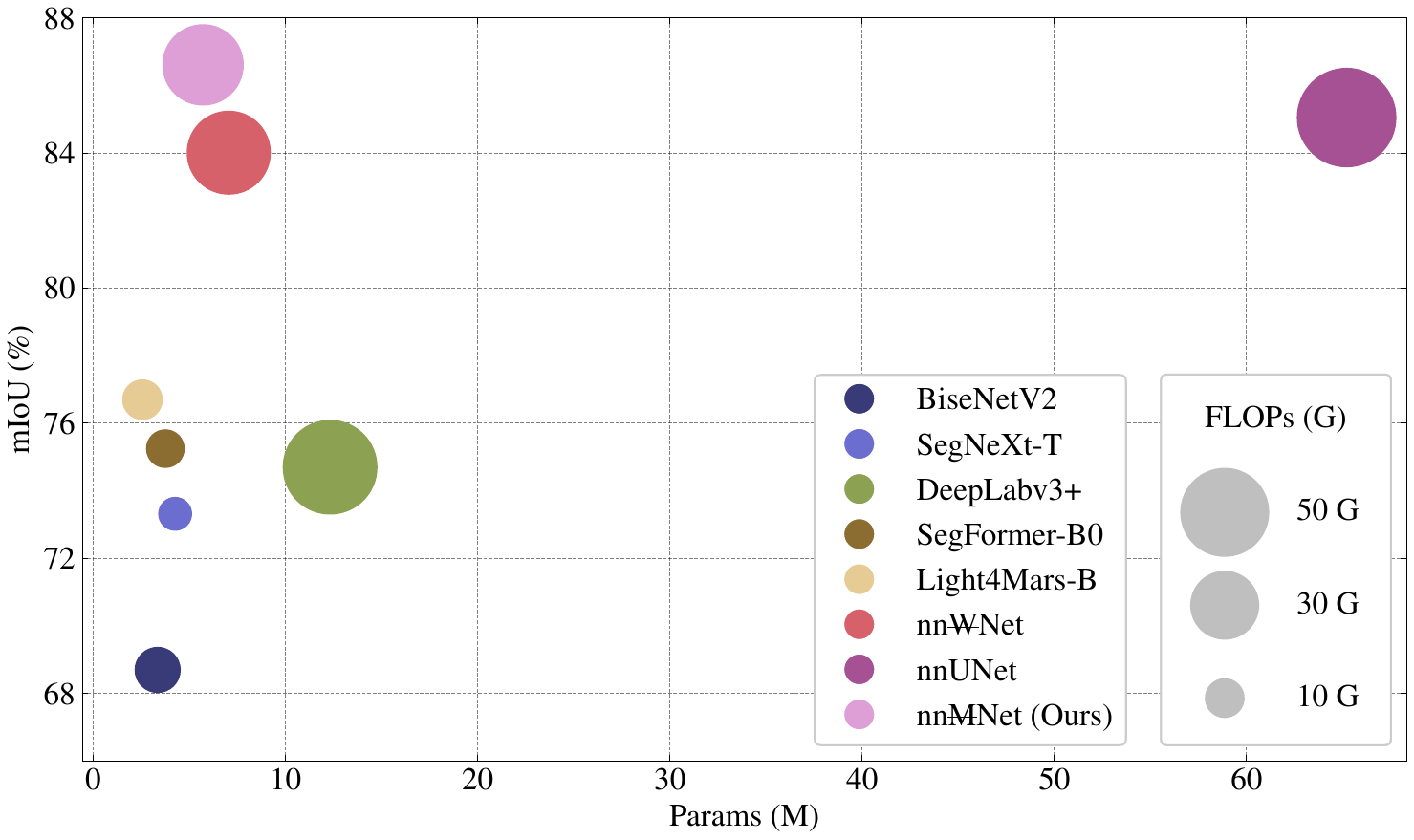}
  \caption{
    Performance comparison on SynMars-TW.
    \nnMNet{} achieves a new state-of-the-art 86.61\% mIoU with low parameters and moderate FLOPs.}
  \label{fig:comparison}
\end{figure}

In this work, we address these challenges by releasing a new baseline model and benchmark. For the model, we begin by examining the limitations of \nnWNet{}~\cite{nnWNet} and then introduce three key components to enhance it. First, we introduce a Linear Attention Block (LAB) to capture global context with linear complexity. Second, we incorporate an Inverted Residual Block (IRB)~\cite{MobileNetV2} to extract local details while reducing computational overhead. Third, we propose a Spatially-Aware Fusion Block (SAFB) to aggregate and refine features from both LAB and IRB in a pixel-wise manner. Together, we present \nnMNet{}, a model that effectively combines the complementary strengths of CNNs and Transformers to obtain superior local and global perception capabilities.

For unified benchmarking, we integrate our proposed model into the nnUNet framework~\cite{nnUNet}, leveraging its end-to-end automated pipeline for standardized training and inference across multiple datasets. Furthermore, we construct a new benchmark comprising three large-scale datasets: SynMars-TW, SynMars-Air, and MarsScapes. Featuring diverse classes with high-quality annotations, these datasets enable a comprehensive evaluation of model performance across varying camera perspectives and data sources.

\nnMNet{} achieves new state-of-the-art results on all datasets in our benchmark, while using fewer parameters and FLOPs than \nnWNet{}. As illustrated in Fig.~\ref{fig:comparison}, it achieves 86.61\% mIoU on SynMars-TW, surpassing all previous methods with comparable model sizes. It also sets new best mIoU scores of 83.25\% and 88.24\% on SynMars-Air and MarsScapes, respectively, demonstrating its robustness and generalizability.

Last but not least, we make our code, models, and benchmark publicly available. To the best of our knowledge, this is the first work to release a reproducible and robust model evaluated on multiple high-quality datasets. We hope our work and observations can provide a solid foundation for advancing research into Martian terrain semantic segmentation.

The main contributions are summarized as follows:
\begin{enumerate}
  \item We introduce Spatially-Aware Fusion Block (SAFB) that performs pixel-level modulation to seamlessly fuse features with diverse characteristics.
  \item We present \nnMNet{}, a novel hybrid CNN-Transformer model for Martian terrain semantic segmentation that effectively models local details and global context.
  \item We curate and standardize three distinct datasets to establish a unified benchmark,
        enabling fair evaluation of the proposed model against previous methods.
  \item We open-source our code, models, and benchmark to provide the field with a new baseline model and a unified benchmark for reproduction and evaluation.
\end{enumerate}

\section{Related Work}
\subsection{CNN-based Methods}
Following the success of FCN~\cite{FCN} and subsequent specialized architectures~\cite{U-Net,SegNet,PSPNet,DeepLab,DeepLabv3+,DANet,SegNeXt}, convolutional neural networks (CNNs) become the dominant paradigm for semantic segmentation. SPOC~\cite{SPOC} pioneers the application of CNNs to Martian imagery by adapting FCN for terrain segmentation using both orbital~\cite{HiRISE} and surface~\cite{MSL} data. Subsequent studies leverage foundational models such as U-Net~\cite{U-Net} and Mask R-CNN~\cite{MaskR-CNN} to segment various Martian landforms and meteorological phenomena~\cite{DS,CraterU-Net,BD}.

To enhance feature representation and make the model focus on important part of the images, attention mechanisms~\cite{SENet,BAM,CBAM,ECA-Net,DANet} have been widely integrated into CNNs to suppress background noise while amplifying foreground details. For instance, HASS~\cite{HASS} introduces a hybrid attention module that jointly models global intra-class coherence and local inter-class distinctions. Additionally, MC-UNet~\cite{MC-UNet} incorporates the Convolutional Block Attention Module (CBAM)~\cite{CBAM} directly into U-Net's skip connections, which emphasize important features across both spatial and channel dimensions.

Alternatively, other studies focus on expanding effective receptive field (ERF)~\cite{ERF} to capture global context. For example, ARS~\cite{ARS} utilizes multi-scale dilated convolutions with varying dilation rates to extract local features from objects of different sizes. Mobile-DeepRFB~\cite{Mobile-DeepRFB} integrates Receptive Field Block (RFB)~\cite{RFB} with a MobileNetV3~\cite{MobileNetV3} backbone within DeepLabV3+~\cite{DeepLabv3+}, enlarging the ERF with fewer computational cost. Similarly, MarsSeg~\cite{MarsSeg} combines a multiscale feature pyramid with strip attention pooling to strengthen the model's high-level semantic understanding.

Although attention mechanisms and dilated convolutions improve CNN's global perception capabilities to some extent, the ERF of these methods remains spatially localized and grows sub-linearly with network depth. Consequently, most of them still struggle to segment continuous, large-scale landforms in highly unstructured Martian environments.

\subsection{Transformer-based Methods}
Recently, Vision Transformers (ViTs) have emerged as a promising alternative to CNNs due to their inherent ability to model global context. However, the quadratic computational complexity of multi-head self-attention (MHSA)~\cite{Transformer} makes it prohibitive for high-resolution images. While DeiT~\cite{DeiT} improves the efficiency through data-efficient training and distillation, the theoretical complexity remains high. The PVT series~\cite{PVT,PVTv2} adopts a downsampling-based approach to improve MHSA, applying spatial reduction to keys and values before computing attention. Meanwhile, the Swin Transformer~\cite{SwinTransformer} introduces shifted windows to restrict attention to local regions while enabling cross-window communication to capture relationships between patches. These two works represent the dominant paradigms for lightweight, high-performance hierarchical ViTs.

In recent years, several studies have adapted these architectures specifically for Mars image segmentation. One line of research focuses on redesigning the attention mechanism. RockFormer~\cite{RockFormer} introduces a PVT~\cite{PVT}-like encoder paired with a feature-refining module to aggregate multi-level features. MarsFormer~\cite{MarsFormer} utilizes MiT~\cite{SegFormer} as backbone and augmented window attention~\cite{SwinTransformer} with strip pooling for feature refinement. Similarly, Light4Mars~\cite{Light4Mars} proposes squeeze window attention, which squeezes queries and keys via strip pooling before MHSA to broaden the receptive field. Furthermore, AirFormer~\cite{AirFormer} decomposes keys and values into different scales and fused them before MHSA to better segment multi-scale objects, and LisseMars~\cite{LisseMars} introduces window movable attention alongside a dynamic polygon convolution module to handle irregular, diverse-sized targets. While most of them integrate convolutions for positional encoding or local feature refinement, their local perception capabilities remain inherently limited due to a lack of inductive biases~\cite{DeiT,CoAtNet}.

Instead of modifying MHSA, another line of research develops hybrid CNN-Transformer models through structural integration. Sequential models, such as SegMarsViT~\cite{SegMarsViT}, interleave lightweight transformer and convolutional blocks, while EDR-TransUnet~\cite{EDR-TransUnet}, RockSeg~\cite{RockSeg} and MarsNet~\cite{MarsNet} replace the deepest convolutional blocks with standard transformer layers. On the other hand, MarsTerrNet~\cite{MarsTerrNet} and Terseg~\cite{TerSeg} adopt dual-branch backbones to extract and fuse features from both architectures. Although these works utilize both CNNs and Transformers, how to fuse their complementary features without losing the delicate balance between local details and global context is yet to be explored.

\subsection{Martian Terrain Semantic Segmentation Datasets}
Advances in deep learning for Martian terrain semantic segmentation depend heavily on the availability of large-scale, high-quality datasets with precise pixel-level annotations. In early days, images of Mars could only be obtained from orbital instruments, such as the High Resolution Imaging Science Experiment (HiRISE)~\cite{HiRISE} and the Context Camera (CTX)~\cite{CTX} aboard NASA's Mars Reconnaissance Orbiter (MRO). The valuable data returned by these instruments laid the foundation for data-driven planetary research from orbit and, in turn, inspired the development of numerous datasets for Martian terrain analysis.

For instance, MC-UNet~\cite{MC-UNet} annotates images from Thermal Emission Imaging System (THEMIS) for impact crater recognition. Similarly, ConeQuest~\cite{ConeQuest} constructs a dataset for Martian cone segmentation using CTX images, with annotations provided by planetary geologists. Furthermore, MarsLS-Net~\cite{MarsLS-Net} integrates CTX images, RGB data, Digital Elevation Model (DEM), slope, and thermal inertia to construct a multi-modal dataset for Martian landslide segmentation. While these orbital datasets remain indispensable for analyzing large-scale geological features, their relatively coarse spatial resolution limits their applicability to finer-scale tasks, such as obstacle avoidance and path planning for Mars rovers.

To better understand high-granularity terrains, researchers have increasingly turned to high-resolution, close-range imagery from Mars rovers. For example, images captured by NASA's Spirit, Opportunity, and Curiosity rovers~\cite{MER, MSL}, along with CNSA's Zhurong rover~\cite{Zhurong}, have been instrumental in this shift. Early efforts in this domain were confined to binary semantic segmentation~\cite{RockFormer,KMF,MarsDataV2, MarsFormer}, where datasets are designed to distinguish foreground obstacles like rocks or stones from the background. More recently, multi-class datasets have become the standard for evaluating holistic scene understanding. These datasets encompass a wider range of terrain types, thus enabling a variety of downstream applications beyond simple obstacle detection. Table~\ref{tab:datasets_summary} summarizes the publicly available Martian terrain semantic segmentation datasets, which provides the diversity and scale required to train complex deep learning models.

\begin{table}[t]
  \caption{Publicly Available Martian Terrain Semantic Segmentation Datasets}
  \label{tab:datasets_summary}
  \centering
  \resizebox{\linewidth}{!}{%
    \begin{tabular}{lllllc}
      \toprule
      Dataset                            & Source   & Type      & Annotation Method & Scale  & Classes \\
      \midrule
      AI4Mars~\cite{AI4Mars}             & MER, MSL & Real      & Crowdsourcing     & 16.39k & 4       \\
      MER-Seg~\cite{Mars-Seg}            & MER      & Real      & Average Labeling  & 1.06k  & 9       \\
      MSL-Seg~\cite{Mars-Seg}            & MSL      & Real      & Average Labeling  & 4.13k  & 9       \\
      \(\text{S}^5\)Mars~\cite{S5Mars}   & MSL      & Real      & Sparse Labeling   & 6k     & 9       \\
      MarsScapes~\cite{MarsScapes, HASS} & MSL      & Real      & Semi-automatic    & 20.8k  & 9       \\
      SynMars~\cite{MarsFormer}          & TW-1     & Synthetic & VisionBlender     & 60k    & 1       \\
      SynMars-TW~\cite{Light4Mars}       & TW-1     & Synthetic & VisionBlender     & 19k    & 8       \\
      SynMars-Air~\cite{AirFormer}       & TW-1     & Synthetic & VisionBlender     & 11.7k  & 8       \\
      \bottomrule
    \end{tabular}
  }
\end{table}

AI4Mars~\cite{AI4Mars} is the first large-scale, open-source dataset for Martian terrain semantic segmentation, comprising 16396 images across four semantic classes. The images come from NASA's Spirit, Opportunity, and Curiosity rovers, as part of the Mars Exploration Rover (MER)~\cite{MER} and Mars Science Laboratory (MSL)~\cite{MSL} missions.  However, because annotations are collected via crowdsourcing, the labels suffer from inconsistencies and boundary inaccuracies. To address these quality issues, Mars-Seg~\cite{Mars-Seg} introduces the MER-Seg and MSL-Seg datasets, which employ twenty-four professional annotators and average their boundary annotations to produce refined ground-truth labels. \(\text{S}^5\)Mars~\cite{S5Mars} adopts sparse labeling, annotating only high human confidence pixels. Both datasets extend the number of classes to nine, enabling more comprehensive evaluation of scene understanding.

While these datasets provide greater scale and semantic diversity, they suffer from inconsistent labeling standards and manual annotation errors that degrade ground-truth quality. MarsScapes~\cite{MarsScapes, HASS} mitigates this through semi-automatic annotation via PixelAnnotationTool~\cite{PixelAnnotationTool}. In contrast, the SynMars series (SynMars~\cite{MarsFormer}, SynMars-TW~\cite{Light4Mars}, and SynMars-Air~\cite{AirFormer}) adopts a fully synthetic approach. Using Blender, they simulate images captured by the Zhurong rover during the Tianwen-1~\cite{Zhurong} mission, and leverage VisionBlender~\cite{VisionBlender} addon to automatically derive synthetic labels from the underlying 3D scenes. In summary, existing datasets are plentiful but remain fragmented, leading most methods to evaluate on a single dataset or a cherry-picked subset. This makes it difficult to assess model generalizability and hinders fair comparison, motivating us to construct a unified benchmark.

\subsection{Local-Global Fusion}
Modern vision models are designed to capture both fine-grained local details and long-range global dependencies. This is typically achieved by combining convolutional and Transformer blocks sequentially~\cite{EarlyConv,LeViT,CoAtNet} or alternately~\cite{MobileViT,CMT,MOAT}. However, these designs force the network to derive global context from localized features or extract local patterns from global representations, often degrading overall performance~\cite{nnWNet}. To address this, parallel architectures process local and global features simultaneously~\cite{nnWNet,iFormer,Conformer,Mobile-Former}. This necessitates dedicated mechanisms to exchange the information and fuse these distinct representations.

Some previous methods adopt simple fusion strategies. \nnWNet{}~\cite{nnWNet} uses channel concatenation to combine features. iFormer~\cite{iFormer} appends a depthwise convolution after concatenation to exchange local information. Conformer~\cite{Conformer} merges branches via element-wise addition with down/up-sampling. Because these methods combine features linearly, they prevent dynamic interaction between local and global representations. Other methods adopt attention mechanisms: Mobile-Former~\cite{Mobile-Former} uses cross-attention to bridge MobileNetV3~\cite{MobileNetV3} and Transformer blocks; Attention U-Net~\cite{AttentionU-Net} applies an attention gate to reweight low-level features with high-level semantic information; FAT~\cite{FAT} leverages element-wise multiplication for bidirectional information exchange.

While these works highlight the value of feature fusion, existing methods fall into two extremes. They are either too simplistic, merging only homogeneous sub-features within blocks, or powerful but come with high computational cost. This motivates us to design a lightweight fusion module tailored for our hybrid CNN-Transformer model.

\section{Method}
\subsection[Limitations in nnWNet]{Limitations in \nnWNet{}}
\nnWNet{}~\cite{nnWNet} is a hybrid CNN-Transformer model originally designed for medical image segmentation. It employs convolutional blocks in the encoders and decoders to capture local details, while using Transformer blocks as bridges to model long-range dependencies. However, this architecture exhibits three primary limitations that prevent it from fully utilizing the strengths of both architectures.

\begin{itemize}
  \item \nnWNet{} replaces MHSA with the PoolFormer~\cite{MetaFormer} block, which sacrifices global receptive field and input-adaptive weighting, resulting in degraded capability for modeling complex scenes such as the Martian surface.
  \item Relying on standard Residual Blocks~\cite{ResNet} leads to higher parameters and FLOPs. This makes the model less viable for systems operating under limited computing resources, such as the onboard systems of Mars rovers.
  \item Feature fusion is achieved through channel concatenation, which forces direct alignment between local and global features. However, this approach hinders performance, as these features focus on objects at different scales.
\end{itemize}

To address these limitations and adapt the model for Martian terrain semantic segmentation, we propose \nnMNet{}, which improves upon \nnWNet{} via three key enhancements. These are detailed in Sec.~\ref{sec:lab},~\ref{sec:irb}, and~\ref{sec:safb}.

\subsection{Overview}
\begin{figure*}[t]
  \centering
  \includegraphics[width=0.8\linewidth]{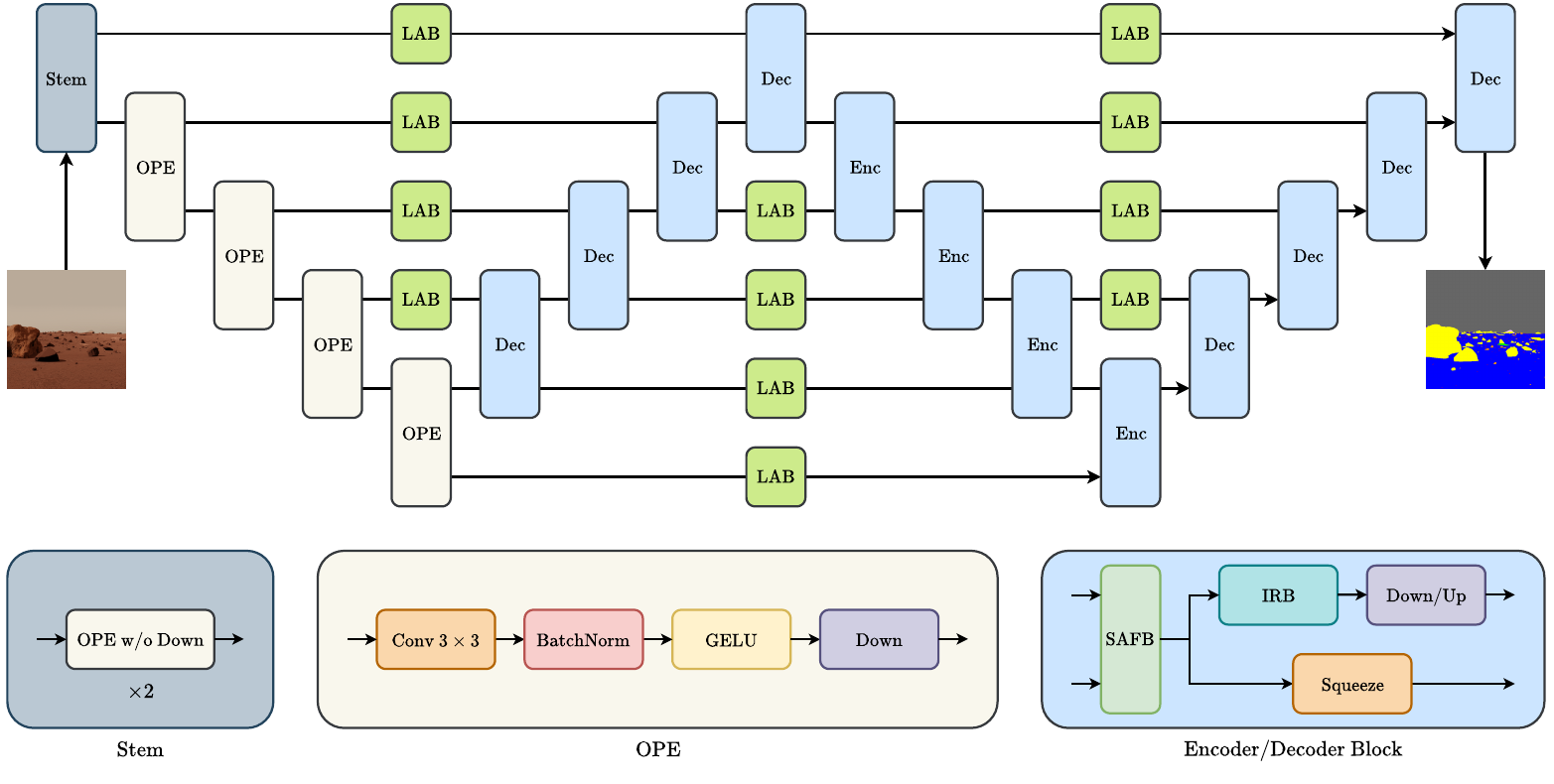}
  \caption{%
    Overview of \nnMNet{}. OPE denotes overlapping patch embedding, LAB denotes Linear Attention Block, IRB denotes Inverted Residual Block, and SAFB denotes Spatially-Aware Fusion Block.
  }
  \label{fig:overview}
\end{figure*}

\begin{figure}[t]
  \centering
  \footnotesize
  \setlength{\tabcolsep}{1pt}
  \begin{tabular}{ccccc}
    \includegraphics[width=0.18\linewidth]{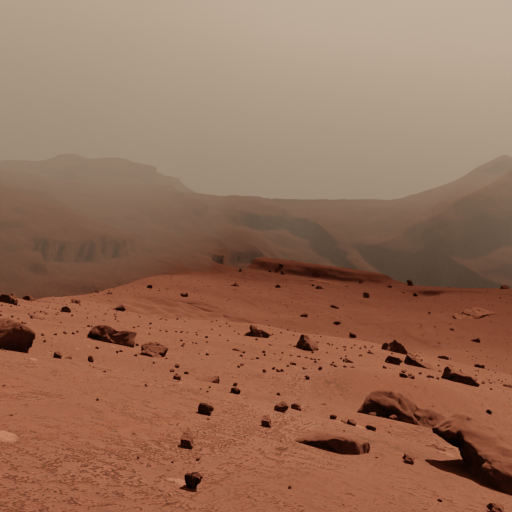}                          &
    \includegraphics[width=0.18\linewidth]{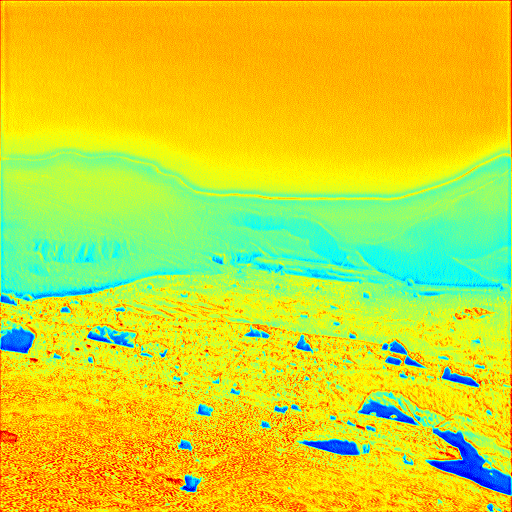} &
    \includegraphics[width=0.18\linewidth]{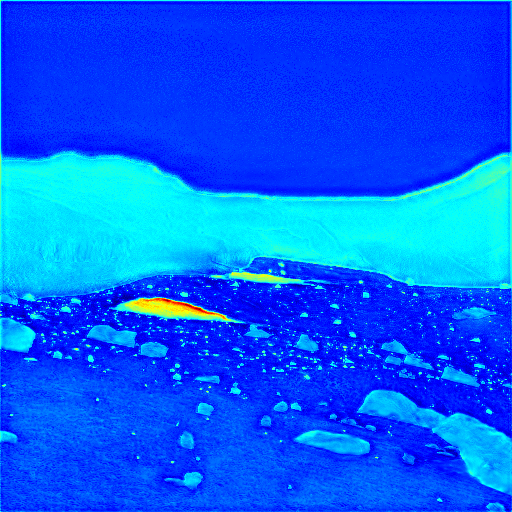} &
    \includegraphics[width=0.18\linewidth]{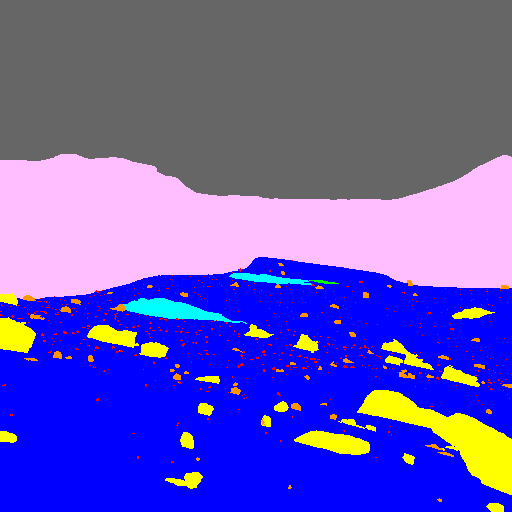}  &
    \includegraphics[width=0.18\linewidth]{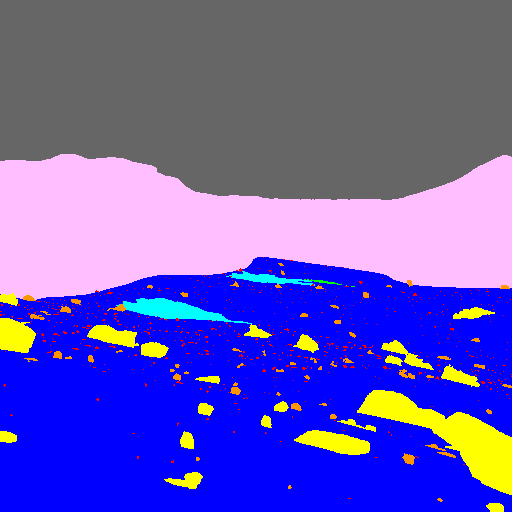}
    \\
    \includegraphics[width=0.18\linewidth]{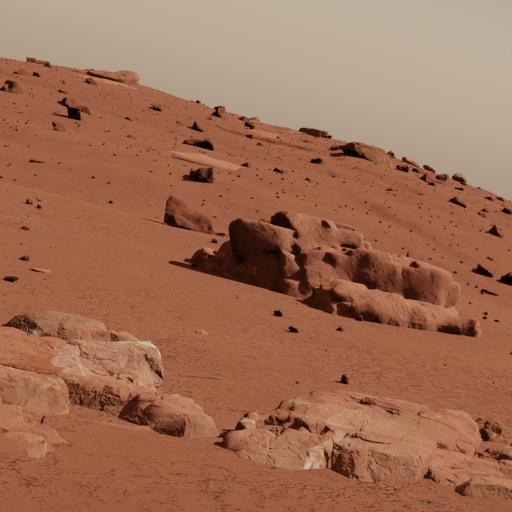}                          &
    \includegraphics[width=0.18\linewidth]{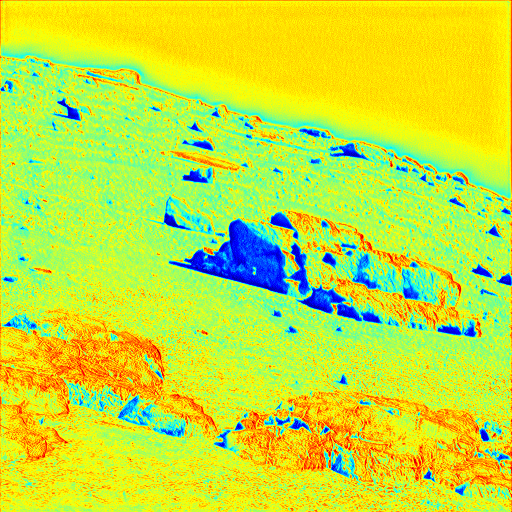} &
    \includegraphics[width=0.18\linewidth]{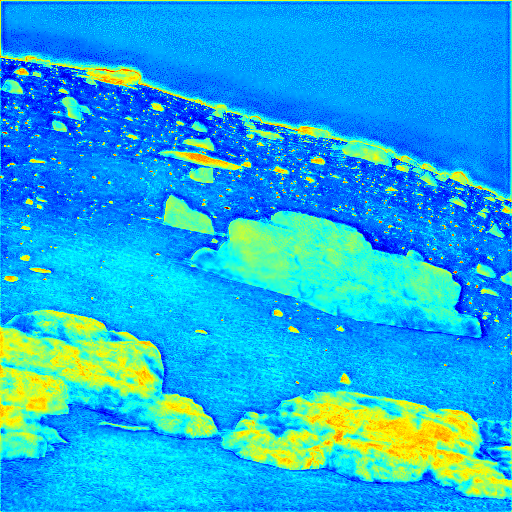} &
    \includegraphics[width=0.18\linewidth]{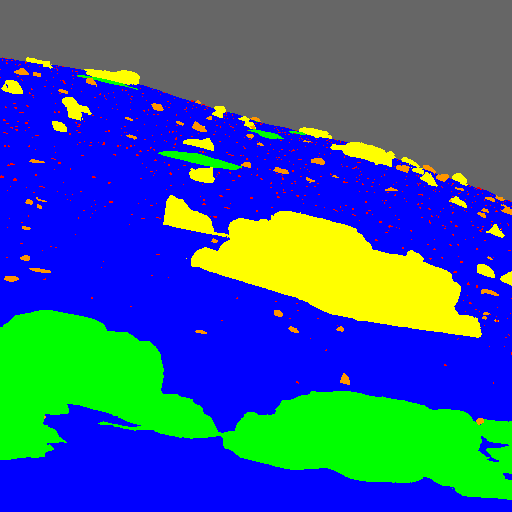}  &
    \includegraphics[width=0.18\linewidth]{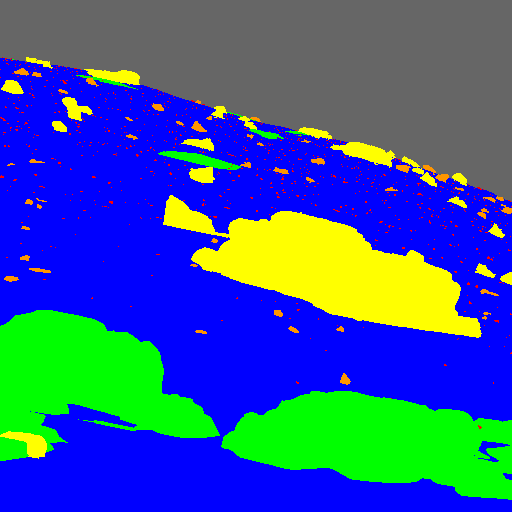}
    \\
    Image                                                                                                                     &
    Coarse                                                                                                                    &
    Fine                                                                                                                      &
    Prediction                                                                                                                &
    Groundtruth
  \end{tabular}
  \caption{%
    Heatmap visualization of the coarse and fine networks on SynMars-TW. They play complementary roles in identifying foreground objects.
  }
  \label{fig:subnetwork_heatmaps}
\end{figure}

As illustrated in Fig.~\ref{fig:overview}, \nnMNet{} comprises two encoder-decoder structures connected via multi-level bridges, forming a cascading architecture. This design decouples the network into coarse network (left) and the fine network (right). As shown in Fig.~\ref{fig:subnetwork_heatmaps}, the coarse network captures broader spatial context and extracts basic features, while the fine network focuses on the foreground objects and refines the final predictions.

The coarse network begins with a convolutional stem without downsampling, which increases optimization stability~\cite{EarlyConv} and preserves details at the original resolution. This stem comprises two overlapping patch embeddings (OPEs); each consists of a 3\(\times\)3 convolution with a stride of 1, batch normalization (BN)~\cite{BatchNormalization}, and a GELU~\cite{GELU} activation function. In subsequent encoder stages, strided OPEs progressively extract local features and project the feature maps into higher-dimensional spaces. Linear Attention Blocks (LABs) bridge the encoder and decoder and preserve global context.

Within each encoder/decoder block, a Spatially-Aware Fusion Block (SAFB) merges features from the preceding stage with those from the bridge. Then, the merged feature is used to generate features for the next stage and the bridge. The feature for the next encoder/decoder is processed by an Inverted Residual Block (IRB)~\cite{MobileNetV2} and a down/up-sampling for reshaping, while the bridge feature is refined by a 3\(\times\)3 convolution to squeeze the channel dimensions. Downsampling is performed with a stride-2 OPE, whereas upsampling is achieved via bilinear interpolation followed by a stride-1 OPE. The fine network shares the same overall architecture as the coarse network but replaces OPEs with encoder-decoder blocks to merge and refine features from the coarse network.

\subsection{Linear Attention Block}
\label{sec:lab}
\begin{figure*}[t]
  \centering
  \includegraphics[width=0.8\linewidth]{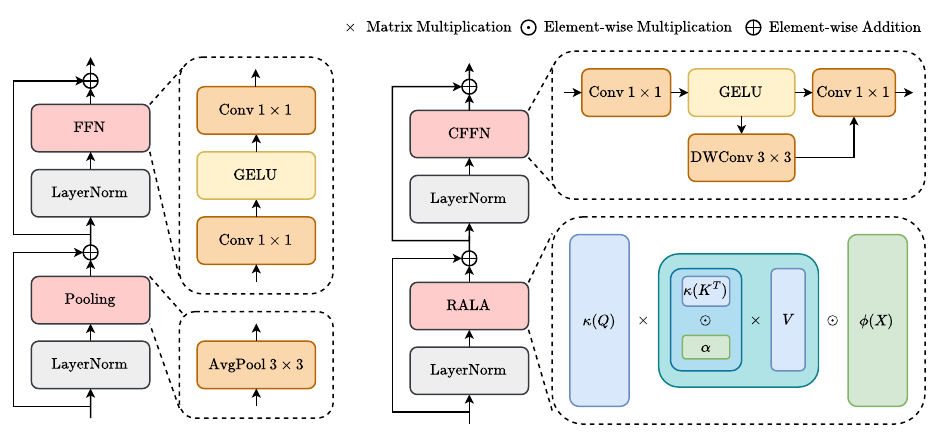}
  \caption{Comparison of the PoolFormer block (left) and Linear Attention Block (LAB) (right).}
  \label{fig:lab}
\end{figure*}

Originally, \nnWNet{} employed the PoolFormer block~\cite{MetaFormer} as the bridge between its encoders and decoders. As illustrated in Fig.~\ref{fig:lab} (left), this block performs token mixing via 3\(\times\)3 average pooling. While computationally efficient, average pooling lacks input-adaptive weighting and a global receptive field, thereby constraining the model's capacity.

To address this, we replace the PoolFormer block with a Linear Attention Block (LAB), shown in Fig.~\ref{fig:lab} (right). Unlike standard self-attention, which suffers from quadratic complexity and a strong local bias that limits global awareness, linear attention enjoys linear complexity and has a more uniform attention distribution~\cite{FlattenTransformer,InLine}. We argue that these properties make it a more suitable complement to convolutions.

\subsubsection{Linear Attention}
We start by revisiting the standard self-attention (Softmax attention)~\cite{Transformer}.
Given an input sequence \(X \in \mathbb{R}^{N \times d}\) consisting of \(N\) tokens with dimension \(d\),
Softmax attention projects \(X\) into queries, keys, and values using learnable weight matrices \(W_Q, W_K, W_V \in \mathbb{R}^{d \times d}\):
\begin{equation}
  Q = XW_Q,\quad K = XW_K,\quad V = XW_V
\end{equation}

The \(i\)-th attention output \(Y_i \in \mathbb{R}^{1 \times d}\) is then computed as:
\begin{equation}
  Y_i = \sum_{j=1}^{N}\frac{\exp\!\left(\frac{Q_i K_j^T}{\sqrt{d_k}}\right)}{\sum_{m=1}^{N}\exp\!\left(\frac{Q_i K_m^T}{\sqrt{d_k}}\right)} V_j
\end{equation}

The term \(\exp(Q_i K_j^T/\sqrt{d_k})\) serves as the similarity function between the \(i\)-th query and the \(j\)-th key. Linear attention approximates this using a kernel function \(\kappa(\cdot)\) such that:
\begin{equation}
  \operatorname{Sim}(Q_i, K_j) = \kappa(Q_i) \kappa(K_j)^T
\end{equation}

After substitution, \(Y_i\) can be rewritten as:
\begin{align}
  Y_i & = \sum_{j=1}^{N}\frac{\operatorname{Sim}(Q_i, K_j)}{\sum_{m=1}^{N}\operatorname{Sim}(Q_i, K_m)} V_j                 \\
      & = \sum_{j=1}^{N}\frac{\kappa(Q_i)\,\kappa(K_j)^T}{\sum_{m=1}^{N}\kappa(Q_i)\,\kappa(K_m)^T} V_j                     \\
      & = \frac{\kappa(Q_i)\bigl(\sum_{j=1}^{N}\kappa(K_j)^T V_j\bigr)}{\kappa(Q_i)\bigl(\sum_{m=1}^{N}\kappa(K_m)^T\bigr)}
\end{align}

By reordering the computation from \((QK^T) V\) to \(Q (K^TV)\), the complexity is reduced from quadratic \(\mathcal{O}(N^2 d)\) to linear \(\mathcal{O}(N d^2)\), as \(d\) is typically much smaller than \(N\).

However, without a Softmax operation to perform nonlinear transform on the weight map, the KV buffer \(\sum_{j=1}^{N}\kappa(K_j)^T V_j\) is used directly to weight the query \(\kappa(Q_i)\). According to the fundamental properties of matrix multiplication, the rank of this KV buffer is bounded by the channel dimension \(d\) (the maximum possible rank of \(K\) and \(V\)). This low-rank bottleneck restricts feature diversity and diminishes the model's representational capacity.

To mitigate this issue, we adopt Rank-Augmented Linear Attention (RALA)~\cite{RALA} as the token mixer within LAB. As illustrated in Fig.~\ref{fig:lab}, RALA introduces two rank augmentation strategies to enhance feature representation and is defined as:
\begin{align}
  Y_i & = \phi(X_i) \odot \left( \frac{\kappa(Q_i) \sum_{j=1}^{N} a_j \, \kappa(K_j)^T V_j} {\kappa(Q_i)\left(\sum_{m=1}^{N}\kappa(K_m)^T\right)} \right), \\
  a_j & = N \cdot \operatorname{softmax}\!\left( Q_g K_j^T \right), \quad
  Q_g = \frac{1}{N}\sum_{i=1}^{N} Q_i .
\end{align}

Here, \(a_j\) serves as the weight coefficient for the \(j\)-th component of the KV buffer. Specifically, \(a_j\) is computed by applying a softmax function to measure the relevance between the \(j\)-th key and a global query \(Q_g\), which is defined as the average of all queries. Applying this scalar weight to the KV buffer augments its rank with negligible computational cost.

Nevertheless, even with this augmented KV buffer, the final output obtained after multiplying by the query \(\kappa(Q_i)\) may still suffer from limited representational capacity.

RALA addresses this by element-wise multiplying the output with the transformed input \(\phi(X_i)\) to raise the rank upper bound. We employ \(\operatorname{ELU}(\cdot) + 1\) as the kernel function \(\kappa(\cdot)\) and a \(1 \times 1\) convolution for the input transformation \(\phi(\cdot)\). Together, the token weighting mechanism and the feature modulation effectively alleviate the low-rank bottleneck of linear attention, enabling richer and more expressive feature representations.

\subsubsection{Positional Encoding}
Although \nnWNet{} omits explicit positional encoding under the assumption that local features inherently carry spatial information, recent work suggests that such spatial cues may partially arise from zero-padding in the average pooling layer~\cite{How}. As RALA lacks zero-padding, this implicit positional information is lost. To address this, we conduct a systematic investigation into how various positional encoding mechanisms affect LAB.

Standard ViTs often employ Absolute Positional Encoding (APE)~\cite{Transformer}, while hierarchical ViTs adopt more flexible formulations to accommodate varying feature map resolutions. For example, Swin Transformer~\cite{SwinTransformer} incorporates Relative Positional Encoding~\cite{RPE} via attention biases. Other architectures use convolution-based approaches, such as Conditional Positional Encoding (CPE)~\cite{CMT, CPVT} and Locally-enhanced Positional Encoding (LePE)~\cite{CSWin}. Alternatively, some designs entirely forgo positional encodings in favor of a Convolutional Feed-Forward Network (CFFN)~\cite{PVTv2, SegFormer, CeiT}. Meanwhile, Rotary Positional Embedding (RoPE)~\cite{RoPE}, which models relative positional relationships via absolute encoding, has been widely adopted in large language models and has shown promise in modern ViTs~\cite{ViT-5}. Together, these methods underscore the importance of positional encoding for ViTs.

\begin{figure*}[t]
  \centering
  \includegraphics[width=0.8\linewidth]{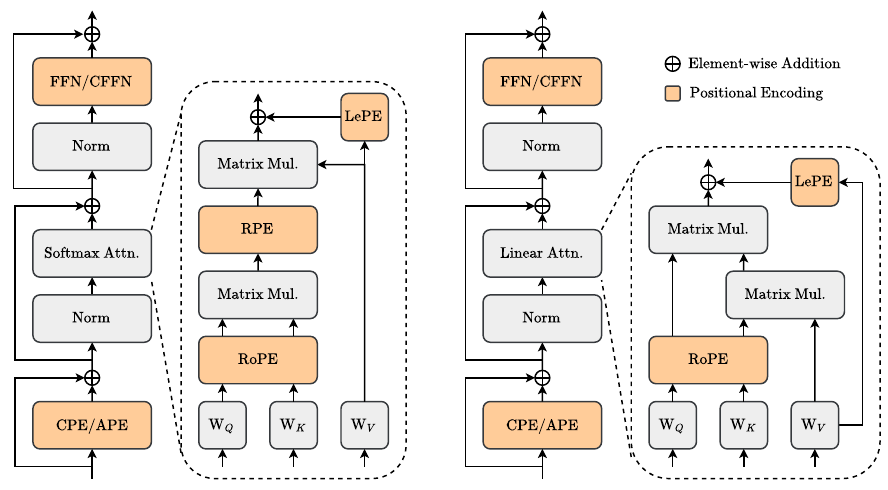}
  \caption{Comparison of positional encodings on self-attention (left) and linear attention (right). RPE does not apply to linear attention.}
  \label{fig:encoding}
\end{figure*}

We evaluate these techniques incrementally, excluding RPE due to its incompatibility with linear attention. As illustrated in Fig.~\ref{fig:encoding}, Softmax attention integrates RPE into the query-key matrix before the Softmax operation, whereas linear attention alters the computation order by computing the key-value product first, thus bypassing the query-key interaction.

For APE, we implement learnable weight vectors with a base size of \(\frac{H}{16} \times \frac{W}{16}\). We apply this encoding before every LAB and use bicubic interpolation to match the size of feature maps across different stages. For CPE, LePE, and CFFN, we follow the common practice and use 3\(\times\)3 depthwise convolution. Finally, we implement 2D RoPE with a frequency base of \(1 \times 10^{-4}\). In our experiments, only RoPE and CFFN improve performance. Therefore, we retain both in the proposed model. The results are detailed in Section~\ref{sec:ablation}.

\subsection{Inverted Residual Block}
\label{sec:irb}
To reduce computational overhead while preserving the model's local perception capabilities, we replace the Residual Blocks~\cite{ResNet} (RBs) in \nnWNet{} with Inverted Residual Blocks (IRBs)~\cite{MobileNetV2}, setting the expansion ratio to 2 to reduce runtime VRAM usage. However, directly substituting RBs with IRBs results in severe performance degradation. We hypothesize that this drop in performance is due to channel dimension mismatches. Specifically, the discrepancy between input and output channel dimensions necessitates more aggressive feature projections before and after expansion, which degrades performance. Additionally, scaling the number of channels before down/up-sampling requires a 1\(\times\)1 convolution within the skip connection, disrupting the pure identity mapping.

To resolve this issue, we increase the number of output channels to match the input dimensions and adjust the dimensions during down/up-sampling instead. This allows us to replace the shortcut projection with a parameter-free identity mapping, which balances the channel expansion and shrinking operations. Following MobileNetV3~\cite{MobileNetV3}, we integrate a Squeeze-and-Excitation (SE) Block~\cite{SENet} after the depthwise convolution to enhance channel-wise feature recalibration.

\subsection{Spatially-Aware Fusion Block}
\label{sec:safb}
In our model, LAB performs channel token mixing after spatial token mixing, deriving cross-channel information from the features with spatial information. Meanwhile, IRB decouples standard convolution into depthwise and pointwise convolutions, each of which partially exploits spatial and channel information, respectively. However, neither module explicitly leverages cross-channel information to enhance spatial details, which motivates us to develop a Spatially-Aware Fusion Block (SAFB) that integrates features extracted by LAB and IRB.

\begin{figure}[t]
  \centering
  \includegraphics[width=\linewidth]{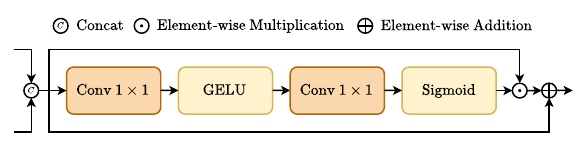}
  \caption{Illustration of the proposed Spatially-Aware Fusion Block.}
  \label{fig:safb}
\end{figure}

As illustrated in Fig.~\ref{fig:safb}, The SAFB at the \(i\)-th stage takes the local feature map \(X_{l} \in \mathbb{R}^{C^{l} \times H_i \times W_i}\) and the global feature map \(X_{g} \in \mathbb{R}^{C^{g} \times H_i \times W_i}\) as inputs. It concatenates these along the channel dimension, then fed it into a two-layer MLP followed by a Sigmoid activation to generate an attention weight map \(A \in \mathbb{R}^{(C^{l} + C^{g}) \times H_i \times W_i}\). To maintain computational efficiency, a bottleneck with a reduction ratio of 4 is introduced in the MLP's hidden layer. The resulting weight map is applied to the concatenated input via element-wise multiplication, yielding an augmented representation that jointly considers both local and global cues. Finally, a residual connection is added to preserve the original information. The overall process is formulated as:
\begin{align}
  Y & = \operatorname{Concat}(X_{l}, X_{g}),                                                                                      \\
  A & = \operatorname{Sigmoid}\bigl(\operatorname{Conv}_{1\times1}(\operatorname{GELU}(\operatorname{Conv}_{1\times1}(Y)))\bigr), \\
  Z & = A \odot Y + Y.
\end{align}

\section{Experiments}
\subsection{Datasets}
\begin{figure}[t]
  \centering
  \footnotesize
  \setlength{\tabcolsep}{1pt}
  \begin{tabular}{@{}ccc@{}}
    \includegraphics[height=0.145\columnwidth]{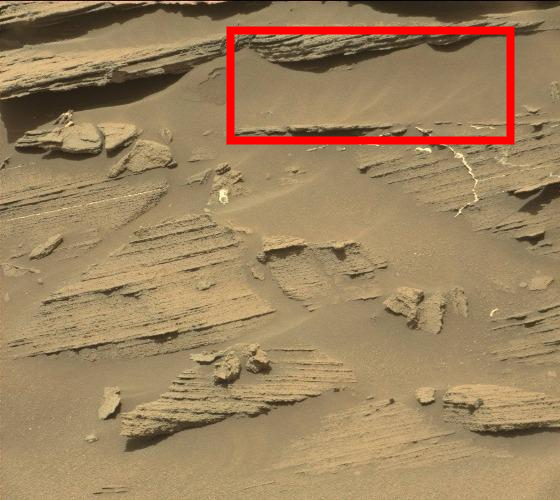}       &
    \includegraphics[height=0.145\columnwidth]{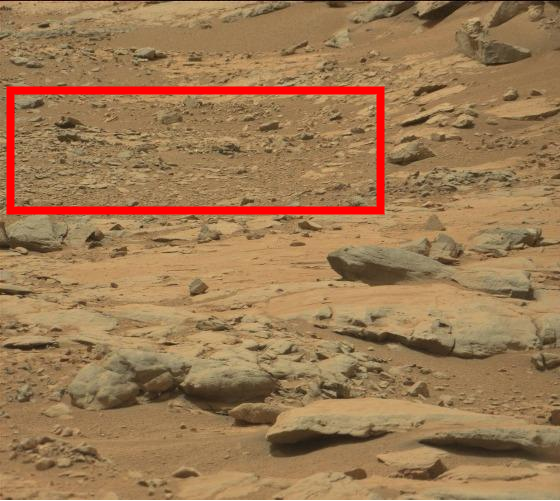}        &
    \includegraphics[height=0.145\columnwidth]{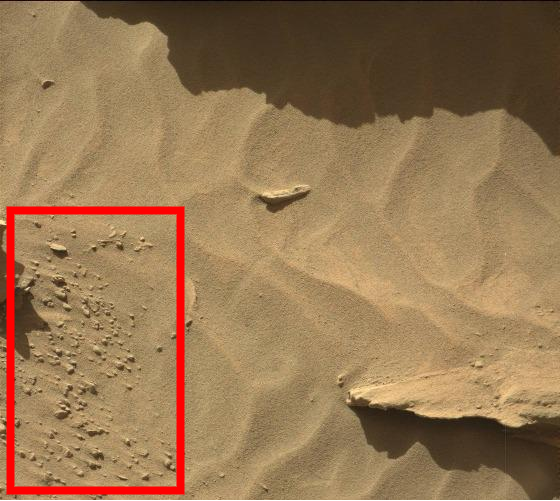}
    \\
    \includegraphics[height=0.145\columnwidth]{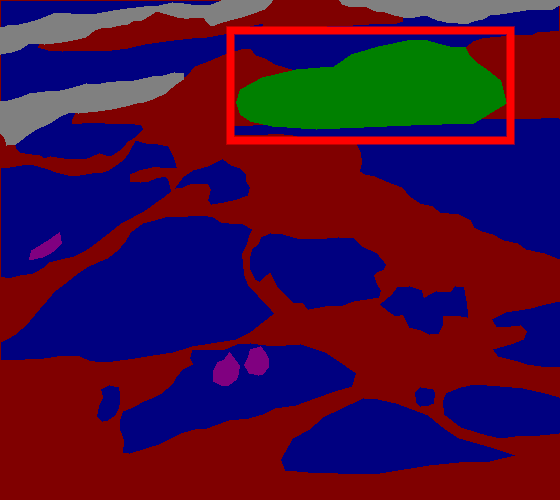} &
    \includegraphics[height=0.145\columnwidth]{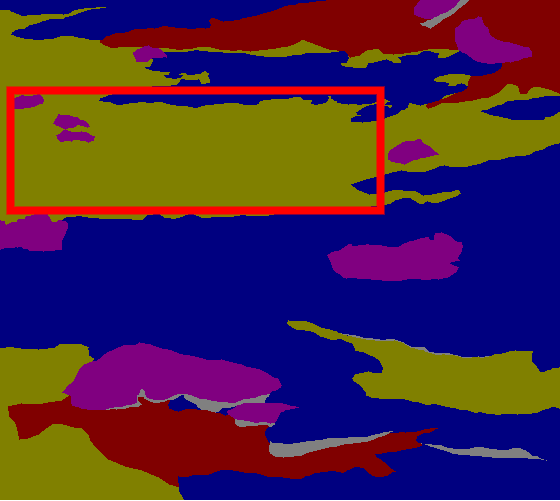}  &
    \includegraphics[height=0.145\columnwidth]{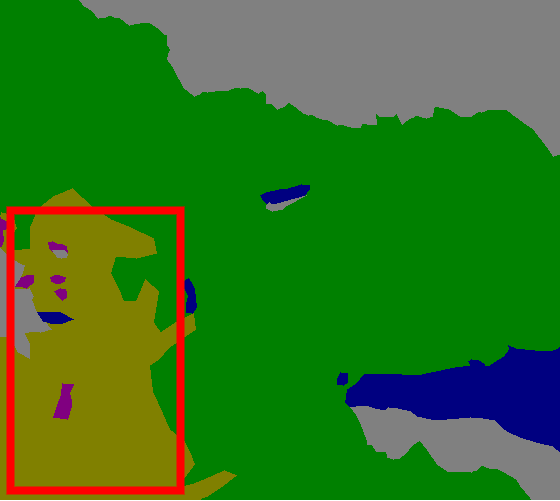}
    \\
    \multicolumn{3}{c}{MSL-Seg}
  \end{tabular}
  \hspace{2pt}
  \begin{tabular}{@{}ccc@{}}
    \includegraphics[height=0.145\columnwidth]{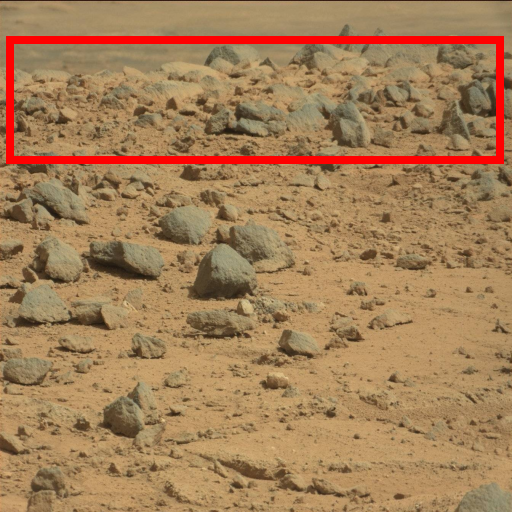}       &
    \includegraphics[height=0.145\columnwidth]{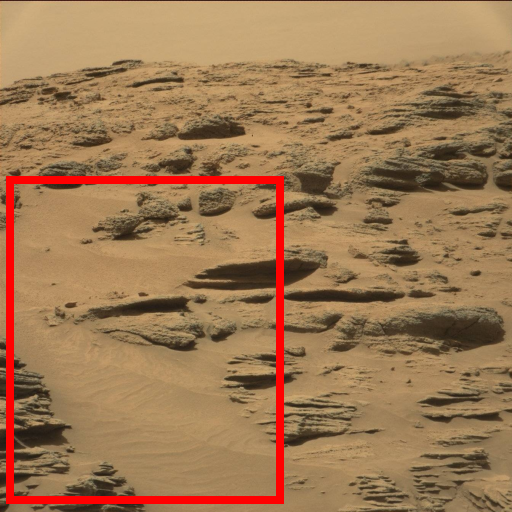}       &
    \includegraphics[height=0.145\columnwidth]{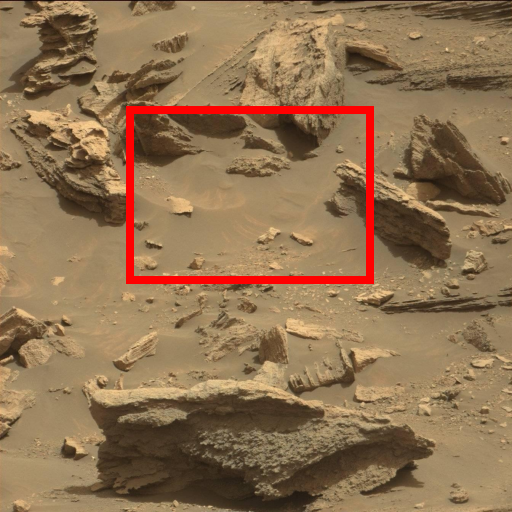}
    \\
    \includegraphics[height=0.145\columnwidth]{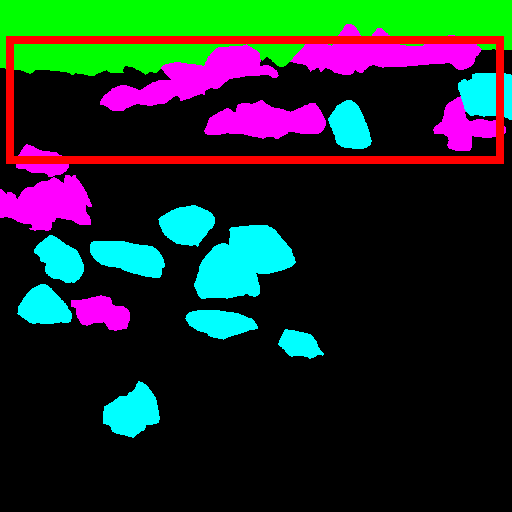} &
    \includegraphics[height=0.145\columnwidth]{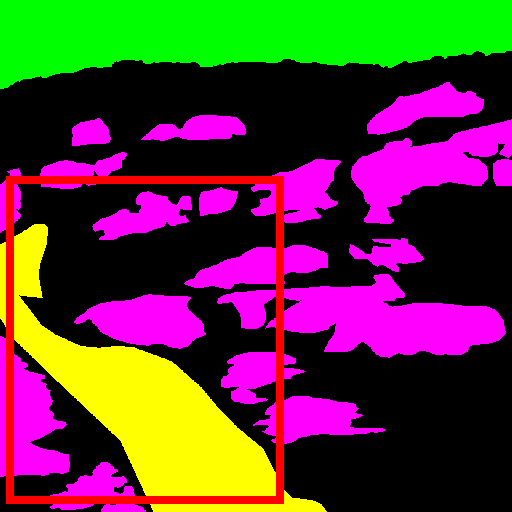} &
    \includegraphics[height=0.145\columnwidth]{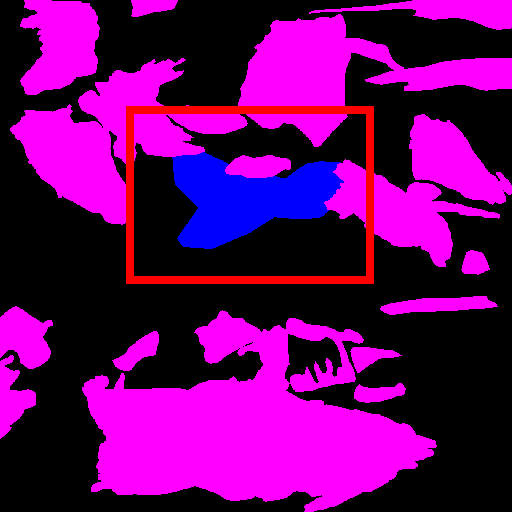}
    \\
    \multicolumn{3}{c}{S5Mars}
  \end{tabular}

  \caption[Examples from MSL-Seg and S5Mars]{
    Examples from MSL-Seg (left) and S5Mars (right).
    The top row shows the input images and the bottom row shows the ground-truth labels.
    \textcolor{red}{Red} boxes highlight regions with ambiguous boundaries or missing labels.
  }
  \label{fig:datasets_visualization}
\end{figure}

We propose a benchmark constructed from a curated selection of existing datasets, as listed in Table~\ref{tab:datasets_summary}. To evaluate dataset quality, we adopt nnUNet~\cite{nnUNet} as a baseline and train it from scratch on each candidate dataset. AI4Mars, MER-Seg, and SynMars are excluded due to input modality (grayscale) and limited class diversity (binary instead of multi-class). As shown in Table~\ref{tab:datasets_results}, nnUNet achieves IoU scores below 50\% for multiple classes on MSL-Seg and \(\text{S}^5\)Mars, resulting in significantly lower mIoU compared to the others. Visual inspection (Fig.~\ref{fig:datasets_visualization}) further reveals frequent ambiguous or mislabeled annotations in these two datasets. Therefore, we evaluate our model exclusively on the three remaining datasets: SynMars-TW, SynMars-Air, and MarsScapes.

The first two are synthetic datasets generated using Blender with the VisionBlender~\cite{VisionBlender} addon.
SynMars-TW simulates images returned from the Zhurong rover and comprises 21,000 images at a 512\(\times\)512 resolution, partitioned into 17,000 training, 2,000 validation, and 2,000 test samples. It defines eight semantic classes: big rock, small rock, gravel, bedrock, ridge, sand, soil, and sky. SynMars-Air extends this to aerial perspectives from the Ingenuity helicopter. It contains 11,700 images of the same resolution and identical classes, divided into 9,400 training, 1,150 validation, and 1,150 test images.

The third dataset is MarsScapes, which consists of real-world images taken by the Curiosity rover and is labeled using PixelAnnotationTool~\cite{PixelAnnotationTool}. It contains 20,802 images at a 256\(\times\)512 resolution, split into 13,002 training, 4,192 validation, and 3,608 test images. MarsScapes comprises nine semantic classes: soil, bedrock, gravel, sand, big rock, ridge, sky, rover, and unknown.

\begin{table}[t]
  \centering
  \caption{nnUNet Results on Martian Terrain Segmentation Datasets}
  \label{tab:datasets_results}
  \resizebox{\linewidth}{!}{%
    \begin{tabular}{lccccc}
      \toprule
      \multirow{2}{*}{Class} & \multicolumn{5}{c}{IoU (\%)}                                                                                         \\
                             & SynMars-TW                   & SynMars-Air            & MarsScapes & MSL-Seg                & \(\text{S}^5\)Mars     \\
      \midrule
      Sky                    & 98.43                        & 99.54                  & 83.06      & —                      & 83.83                  \\
      Ridge                  & 93.49                        & 81.86                  & 73.59      & —                      & 82.38                  \\
      Soil / Martian Soil    & 98.79                        & 99.11                  & 86.89      & \textcolor{red}{17.24} & \textcolor{red}{30.39} \\
      Sand / Sands           & 77.73                        & 65.33                  & 73.91      & \textcolor{red}{41.00} & \textcolor{red}{37.69} \\
      Bedrock                & 68.53                        & 70.18                  & 75.13      & 59.11                  & 57.87                  \\
      Gravel                 & 51.65                        & \textcolor{red}{37.88} & 72.42      & 58.73                  & —                      \\
      Big rock / Rock(s)     & 92.32                        & 85.35                  & 66.63      & \textcolor{red}{32.69} & \textcolor{red}{22.14} \\
      Small rock             & 72.39                        & 72.19                  & —          & —                      & —                      \\
      Rover                  & —                            & —                      & 59.50      & —                      & 71.80                  \\
      Tracks / Trace         & —                            & —                      & —          & \textcolor{red}{38.32} & \textcolor{red}{20.06} \\
      Shadows                & —                            & —                      & —          & \textcolor{red}{37.44} & —                      \\
      Hole                   & —                            & —                      & —          & —                      & \textcolor{red}{41.46} \\
      Background / Unlabeled & —                            & —                      & 67.24      & 55.33                  & —                      \\
      Unknown                & —                            & —                      & —          & 70.36                  & —                      \\
      \midrule
      mIoU (\%)              & 81.66                        & 76.43                  & 73.15      & \textcolor{red}{45.58} & \textcolor{red}{49.74} \\
      \bottomrule
    \end{tabular}
  }
\end{table}

\subsection{Implementation details}
We conduct all experiments on two NVIDIA GeForce RTX 4090 GPUs using nnUNet framework~\cite{nnUNet}. Following the framework's default configuration, the model is trained from scratch for 1000 epochs (250K iterations). For the ablation studies, we limit training to 200 epochs (50K iterations) due to computational constraints. To stabilize training of LAB, we adopt the AdamW~\cite{AdamW} optimizer with an initial learning rate of 4e-4 and a weight decay of 0.05. In contrast, nnUNet and \nnWNet{} retain their default SGD optimizers as per their original implementations. The training configurations are automatically configured by the framework. We evaluate semantic segmentation performance using mean Intersection over Union (mIoU), with inference speed reported in frames per second (FPS) on an NVIDIA GeForce RTX 4090 GPU.

\subsection{Comparison with State-of-the-Art Methods}
In this section, we compare our method with state-of-the-art methods. As our model is built upon the nnUNet framework, we first compare with nnUNet and \nnWNet{}, which are originally designed for medical image semantic segmentation. We then evaluate against general vision models of similar parameters, including BiseNetV2~\cite{BiseNetV2}, DeepLabv3+~\cite{DeepLabv3+}, SegFormer-B0~\cite{SegFormer}, and SegNeXt-T~\cite{SegNeXt}. Finally, we compare with the state-of-the-art method in Martian terrain semantic segmentation: Light4Mars-B~\cite{Light4Mars}, the only model that is publicly available and implemented under a widely adopted framework.

\subsubsection{Results on SynMars-TW}
We compare our \nnMNet{} with existing methods on SynMars-TW. As shown in Table~\ref{tab:synmars_tw}, \nnMNet{} achieves a new state-of-the-art mIoU of 86.61\% and performs best across multiple categories. It outperforms Light4Mars-B by 36.39\% on Small Rock and 58.61\% on Gravel. Against \nnWNet{}, it improves mIoU by 2.60\% (84.01\% \(\rightarrow\) 86.61\%) while reducing parameters by 19.01\% (7.05M \(\rightarrow\) 5.71M) and FLOPs by 6.42\% (42.54G \(\rightarrow\) 39.81G). Compared to nnUNet, it achieves superior results with only 8.75\% of its parameters and 66.13\% of its FLOPs, demonstrating the efficiency of our hybrid architecture.

\begin{table*}[t]
  \centering
  \caption{Comparison with state-of-the-art methods on SynMars-TW}
  \label{tab:synmars_tw}
  \resizebox{\linewidth}{!}{%
    \begin{tabular}{l|cccccccccccc}
      \toprule
      Method                         & Big rock       & Small rock     & Gravel         & Bedrock        & Ridge          & Sand           & Soil           & Sky            & mIoU (\%)      & Params (M) & FLOPs (G) & FPS    \\
      \midrule
      BiseNetV2~\cite{BiseNetV2}     & 83.58          & 22.09          & 0.01           & 84.43          & 95.95          & 68.02          & 96.09          & 99.41          & 68.70          & 3.35       & 12.31     & 501.13 \\
      DeepLabv3+~\cite{DeepLabv3+}   & 88.80          & 43.98          & 1.30           & 89.46          & \textbf{96.95} & 80.25          & 97.33          & \textbf{99.52} & 74.70          & 12.32      & 54.27     & 192.78 \\
      SegFormer-B0~\cite{SegFormer}  & 89.74          & 42.86          & 1.41           & 91.06          & 95.56          & 84.43          & 97.65          & 99.30          & 75.25          & 3.75       & 8.50      & 302.28 \\
      SegNeXt-T~\cite{SegNeXt}       & 89.24          & 31.43          & 0.23           & 89.41          & 95.88          & 83.83          & 97.16          & 99.36          & 73.32          & 4.26       & 6.45      & 221.09 \\
      Light4Mars-B~\cite{Light4Mars} & 89.81          & 44.37          & 0.09           & \textbf{91.70} & 96.65          & \textbf{84.78} & 97.84          & 99.47          & 76.70          & 2.56       & 9.42      & 160.51 \\
      nnUNet~\cite{nnUNet}           & 94.31          & 78.10          & 56.61          & 76.44          & 94.77          & 81.73          & 99.08          & 99.36          & 85.05          & 65.22      & 60.20     & 274.69 \\
      \nnWNet{}~\cite{nnWNet}        & 94.26          & 77.32          & 54.81          & 74.22          & 93.15          & 80.79          & 98.98          & 98.53          & 84.01          & 7.05       & 42.54     & 92.58  \\
      \textbf{\nnMNet{}} (Ours)      & \textbf{95.48} & \textbf{80.76} & \textbf{58.70} & 79.83          & 95.44          & 84.22          & \textbf{99.17} & 99.26          & \textbf{86.61} & 5.71       & 39.81     & 63.33  \\
      \bottomrule
    \end{tabular}
  }
\end{table*}

\subsubsection{Results on SynMars-Air}
To evaluate generalizability to aerial perspectives, we compare the proposed method with existing methods on SynMars-Air. As shown in Table~\ref{tab:synmars_air}, \nnMNet{} achieves a new best mIoU of 83.25\%, surpassing the \nnWNet{} by 2.64\% (80.61\% \(\rightarrow\) 83.25\%). While our method performs slightly worse than non-nnUNet-based architectures on medium-scale classes such as Bedrock, Ridge, and Sand, it provides more balanced performance and superior overall segmentation accuracy across all categories.

\begin{table*}[t]
  \centering
  \caption{Comparison with state-of-the-art methods on SynMars-Air}
  \label{tab:synmars_air}
  \resizebox{\linewidth}{!}{%
    \begin{tabular}{l|cccccccccccc}
      \toprule
      Method                         & Big rock       & Small rock     & Gravel         & Bedrock        & Ridge          & Sand           & Soil           & Sky            & mIoU (\%)      & Params (M) & FLOPs (G) & FPS    \\
      \midrule
      BiseNetV2~\cite{BiseNetV2}     & 84.74          & 29.14          & 0.00           & 90.00          & 98.27          & 96.63          & 98.66          & 99.77          & 74.65          & 3.35       & 12.31     & 501.13 \\
      DeepLabv3+~\cite{DeepLabv3+}   & 87.86          & 43.31          & 0.34           & 92.55          & 98.71          & 95.54          & 99.03          & 99.81          & 77.15          & 12.32      & 54.27     & 192.78 \\
      SegFormer-B0~\cite{SegFormer}  & 90.56          & 52.67          & 1.09           & 94.33          & 98.83          & \textbf{98.24} & 99.26          & 99.75          & 79.34          & 3.75       & 8.50      & 302.28 \\
      SegNeXt-T~\cite{SegNeXt}       & \textbf{91.36} & 43.62          & 0.07           & 94.02          & \textbf{98.97} & 98.21          & 99.10          & 99.83          & 78.15          & 4.26       & 6.45      & 221.09 \\
      Light4Mars-B~\cite{Light4Mars} & 89.98          & 49.81          & 5.02           & \textbf{94.38} & 98.94          & 97.95          & 99.26          & 99.83          & 79.40          & 2.56       & 9.42      & 160.51 \\
      nnUNet~\cite{nnUNet}           & 89.40          & 77.94          & 43.04          & 77.62          & 87.58          & 74.39          & 99.31          & 98.01          & 80.91          & 65.22      & 60.20     & 274.69 \\
      \nnWNet{}~\cite{nnWNet}        & 88.04          & 76.86          & 41.86          & 75.45          & 90.94          & 75.37          & 99.30          & 97.05          & 80.61          & 7.05       & 42.54     & 92.58  \\
      \textbf{\nnMNet{}} (Ours)      & 90.66          & \textbf{80.17} & \textbf{45.68} & 78.61          & 93.93          & 77.62          & \textbf{99.35} & \textbf{99.96} & \textbf{83.25} & 5.71       & 39.81     & 63.33  \\
      \bottomrule
    \end{tabular}
  }
\end{table*}

\subsection{Results on MarsScapes}
We evaluate the proposed model on real-world data using MarsScapes. As shown in Table~\ref{tab:marsscapes}, \nnMNet{} achieves a new best mIoU of 88.24\%. Although the margin is narrow (+0.92\%), our method consistently outperforms \nnWNet{} and secures top performance in five out of nine terrain categories, demonstrating its effectiveness in real-world scenes.

\begin{table*}[t]
  \centering
  \caption{Comparison with state-of-the-art methods on MarsScapes}
  \label{tab:marsscapes}
  \resizebox{\linewidth}{!}{%
    \begin{tabular}{l|ccccccccccccc}
      \toprule
      Method                         & Soil           & Bedrock        & Gravel         & Sand           & Big Rock       & Ridge          & Sky            & Rover          & Unlabeled      & mIoU (\%)      & Params (M) & FLOPs (G) & FPS    \\
      \midrule
      BiseNetV2~\cite{BiseNetV2}     & 89.30          & 84.20          & 74.16          & 81.16          & 63.43          & 88.68          & 95.47          & 77.78          & 79.22          & 81.49          & 3.35       & 6.16      & 623.24 \\
      DeepLabv3+~\cite{DeepLabv3+}   & 85.94          & 80.52          & 66.04          & 78.15          & 57.36          & 83.70          & 91.68          & 63.75          & 72.03          & 75.47          & 12.32      & 27.14     & 300.22 \\
      SegFormer-B0~\cite{SegFormer}  & 91.12          & 86.59          & 78.25          & 84.16          & 67.50          & 92.46          & 96.61          & 88.64          & \textbf{84.25} & 85.51          & 3.72       & 3.70      & 395.43 \\
      SegNeXt-T~\cite{SegNeXt}       & 91.71          & 87.21          & 79.81          & 84.18          & 67.98          & \textbf{93.00} & \textbf{96.92} & 88.11          & 83.87          & 85.87          & 4.26       & 3.23      & 252.14 \\
      Light4Mars-B~\cite{Light4Mars} & 90.78          & 86.59          & 77.11          & 83.64          & 66.24          & 91.35          & 95.97          & 83.68          & 81.96          & 84.15          & 2.56       & 4.82      & 282.06 \\
      nnUNet~\cite{nnUNet}           & 93.28          & 87.45          & 85.24          & 84.67          & 83.19          & 80.73          & 90.13          & \textbf{97.85} & 80.00          & 86.95          & 47.64      & 29.84     & 387.84 \\
      \nnWNet{}~\cite{nnWNet}        & 93.63          & 87.96          & 85.38          & 85.03          & 82.23          & 82.81          & 91.26          & 97.18          & 80.37          & 87.32          & 7.05       & 21.27     & 165.88 \\
      \textbf{\nnMNet{}} (Ours)      & \textbf{93.85} & \textbf{88.25} & \textbf{85.80} & \textbf{85.98} & \textbf{84.28} & 85.19          & 91.73          & 97.80          & 81.31          & \textbf{88.24} & 5.71       & 19.91     & 93.45  \\
      \bottomrule
    \end{tabular}
  }
\end{table*}

\subsection{Ablation Study}
\label{sec:ablation}
\subsubsection{Impact of Component Designs}
To evaluate the contribution of each component in \nnMNet{}, we add them incrementally and conduct experiments on SynMars-TW, as summarized in Table~\ref{tab:ablation}. We begin by investigating the design of the proposed LAB. Replacing average pooling with RALA improves mIoU by 0.45\% with only a marginal increase in parameters and FLOPs. Among the positional encoding mechanisms integrated into LAB, RoPE proves to be the most effective. However, adding APE, CPE, or LePE yields no further improvements. We hypothesize that RoPE already supplies sufficient spatial information, and that additional positional encodings lead to interference. Although adding a CFFN only provides a slight accuracy increase (+0.05\%), we retain it for implicit local information and training stability.

We further ablate the design of the improved IRB. Using IRB as a drop-in replacement for RB significantly reduces parameters and FLOPs but causes a noticeable performance drop. To mitigate this, we increase the number of output channels to preserve a pure identity mapping and expand the hidden layer, achieving a modest performance boost (+0.18\%) while maintaining lower computational cost compared to the RB-based version. Finally, we integrate the SE Block to facilitate channel-wise information exchange and the proposed SAFB to better fuse features from LAB and IRB. These two modules yield similar performance boosts (+0.15\% and +0.12\%, respectively) with acceptable cost. Together, our proposed designs bring a total +1.38\% mIoU improvement over over the baseline \nnWNet{}.

\begin{table}[t]
  \centering
  \caption{Ablation of \nnMNet{} components}
  \label{tab:ablation}
  \begin{tabular}{l|ccc}
    \toprule
    Method                                                  & mIoU (\%)                                   & Params (M) & FLOPs (G) \\
    \midrule
    \nnWNet{}~\cite{nnWNet}                                 & 81.17 (\(\uparrow\)0.00)                    & 7.05       & 42.54     \\
    Pooling \(\rightarrow\) RALA~\cite{RALA}                & 81.62 (\(\uparrow\)0.45)                    & 7.21       & 43.91     \\
    + RoPE~\cite{RoPE}                                      & 82.05 (\(\uparrow\)0.43)                    & 7.21       & 43.91     \\
    + APE~\cite{Transformer}                                & 81.64 \textcolor{red}{(\(\downarrow\)0.41)} & 9.18       & 43.91     \\
    + CPE~\cite{CMT, CPVT}                                  & 81.79 \textcolor{red}{(\(\downarrow\)0.26)} & 7.22       & 44.00     \\
    + LePE~\cite{CSWin}                                     & 81.78 \textcolor{red}{(\(\downarrow\)0.27)} & 7.22       & 44.00     \\
    + CFFN~\cite{PVTv2, SegFormer, CeiT}                    & 82.10 (\(\uparrow\)0.05)                    & 7.23       & 44.26     \\
    RB~\cite{ResNet} \(\rightarrow\) IRB~\cite{MobileNetV2} & 81.38 (\(\downarrow\)0.72)                  & 4.06       & 30.78     \\
    Keep identity mapping                                   & 82.28 (\(\uparrow\)0.90)                    & 4.84       & 37.71     \\
    + SE Block~\cite{SENet}                                 & 82.43 (\(\uparrow\)0.15)                    & 5.56       & 37.71     \\
    + SAFB                                                  & 82.55 (\(\uparrow\)0.12)                    & 5.71       & 39.81     \\
    \bottomrule
  \end{tabular}
\end{table}

\subsubsection{Choice of Token Mixer}
To validate the linear attention design in our hybrid CNN-Transformer architecture, we replace the token mixer within LAB with two alternatives: the Efficient Self-Attention from SegFormer~\cite{SegFormer} and the average pooling from PoolFormer~\cite{MetaFormer}. In Table~\ref{tab:token_mixer}, RALA~\cite{RALA} achieves the highest mIoU (82.55\%) while maintaining a modest computational budget. The performance drop observed with Efficient Self-Attention indicates that its spatial-reduction mechanism can impair model's global perception capabilities. Average pooling yields better accuracy than Efficient Self-Attention but still falls short of RALA, confirming that input-adaptive weighting and global receptive field are necessary for increasing the model capacity.

\begin{table}[t]
  \centering
  \caption{Ablation of the token mixer within LAB}
  \label{tab:token_mixer}
  \begin{tabular}{l|ccc}
    \toprule
    token mixer                               & mIoU (\%) & Params (M) & FLOPs (G) \\
    \midrule
    RALA~\cite{RALA}                          & 82.55     & 5.71       & 39.81     \\
    Efficient Self-Attention~\cite{SegFormer} & 81.99     & 6.47       & 44.11     \\
    Average Pooling~\cite{MetaFormer}         & 82.13     & 5.55       & 38.46     \\
    \bottomrule
  \end{tabular}
\end{table}

\subsubsection{Influence of Kernel Size}
Modern CNNs often adopt large kernels~\cite{VAN,RepLKNet,SLaK,LSKA,ConvNext,SegNeXt} to capture more global information. We examine how the kernel size of the depthwise convolution within IRBs affects our model. As Table~\ref{tab:kernel_size} shows, enlarging the kernel from 3\(\times\)3 to 7\(\times\)7 consistently degrades performance. We attribute this to the fact that LAB already provide sufficient global information and effectively enlarge the ERF. Under these conditions, larger kernels introduce redundant noise rather than additional global context, whereas smaller kernels allow the model to focus more on extracting fine-grained local details.

\begin{table}[t]
  \centering
  \caption{Ablation of IRB kernel size}
  \label{tab:kernel_size}
  \begin{tabular}{l|ccc}
    \toprule
    kernel size  & mIoU (\%) & Params (M) & FLOPs (G) \\
    \midrule
    3\(\times\)3 & 82.55     & 5.71       & 39.81     \\
    5\(\times\)5 & 82.36     & 5.77       & 40.56     \\
    7\(\times\)7 & 82.33     & 5.85       & 41.69     \\
    \bottomrule
  \end{tabular}
\end{table}

\subsubsection{Ablation of SAFB Reduction Ratio}
We conduct an ablation study on the reduction ratio of the MLP in the SAFB. The results, presented in Table~\ref{tab:safb}, show that a reduction ratio of 0.25 achieves the best trade-off between model performance and computational complexity.

\begin{table}[t]
  \centering
  \caption{Ablation of SAFB Reduction Ratio}
  \label{tab:safb}
  \begin{tabular}{l|ccc}
    \toprule
    reduction ratio & mIoU (\%) & Params (M) & FLOPs (G) \\
    \midrule
    1               & 82.57     & 6.16       & 42.57     \\
    0.5             & 82.54     & 5.86       & 40.73     \\
    0.25            & 82.55     & 5.71       & 39.81     \\
    0.125           & 82.38     & 5.64       & 39.95     \\
    w/o SAFB        & 82.43     & 5.56       & 37.71     \\
    \bottomrule
  \end{tabular}
\end{table}

\subsubsection{Impact of Single vs. Two-Stage Architectures}
We evaluate the impact of the network architecture by comparing single-stage and two-stage encoder-decoder variants, with results presented in Table~\ref{tab:encoder_decoder}. The single-stage variant, \nnMNetPlus{}, is constructed by extracting and scaling the fine network to a comparable size. Notably, it achieves 58.33\% speedup compared to the two-stage variant and is 8.31\% faster than \nnWNet{}, When trained for 50K iterations, the single-stage variant yields better results than the two-stage variant. However, when scaled to 250K iterations, the two-stage variant outperforms the single-stage variant by 0.49\%. We hypothesize that the single-stage design is simpler to optimize, whereas the two-stage design converges more slowly but has a larger capacity for peak performance. Although the two-stage variant achieves the highest accuracy and is chosen as our final model, the single-stage variant delivers faster inference, which is preferable for real-world deployment.

\begin{table}[t]
  \centering
  \caption{Ablation of Single versus Two-Stage Architectures}
  \label{tab:encoder_decoder}
  \resizebox{\columnwidth}{!}{%
    \begin{tabular}{l|ccccc}
      \toprule
      Model         & Iterations & mIoU (\%) & Params (M) & FLOPs (G) & FPS                    \\
      \midrule
      \nnWNet{}     & 50K        & 80.02     & 7.05       & 42.54     & 92.58                  \\
      \nnMNet{}     & 50K        & 82.55     & 5.71       & 39.81     & 63.33                  \\
      \nnMNetPlus{} & 50K        & 82.66     & 4.23       & 39.32     & 100.27                 \\
      \midrule
      \nnWNet{}     & 250K       & 84.01     & 7.05       & 42.54     & 92.58                  \\
      \nnMNet{}     & 250K       & 86.61     & 5.71       & 39.81     & 63.33                  \\
      \nnMNetPlus{} & 250K       & 86.12     & 4.23       & 39.32     & 100.27                 \\
      \bottomrule
      \multicolumn{6}{@{}l@{}}{\footnotesize \(^{\dagger}\) denotes the single-stage variant.} \\
    \end{tabular}
  }
\end{table}

\subsection{Qualitative Results}
\subsubsection{Results on the benchmark}
\begin{figure*}[t]
  \centering
  \footnotesize
  \setlength{\tabcolsep}{1pt}
  \begin{tabular}{ccccccc}
    \includegraphics[width=0.13\linewidth]{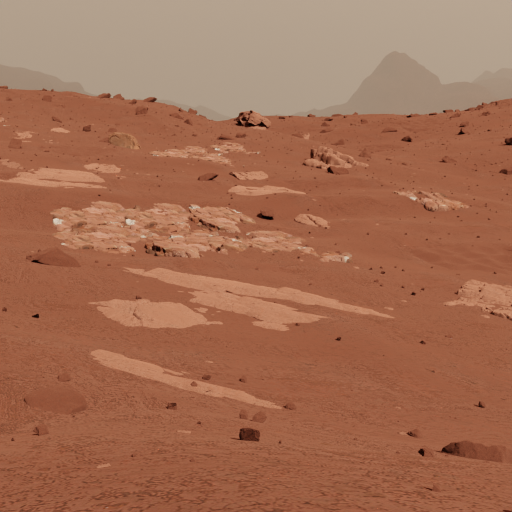}                         &
    \includegraphics[width=0.13\linewidth]{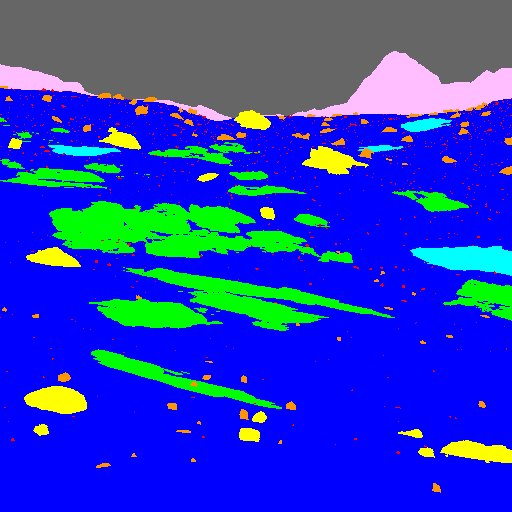}                   &
    \includegraphics[width=0.13\linewidth]{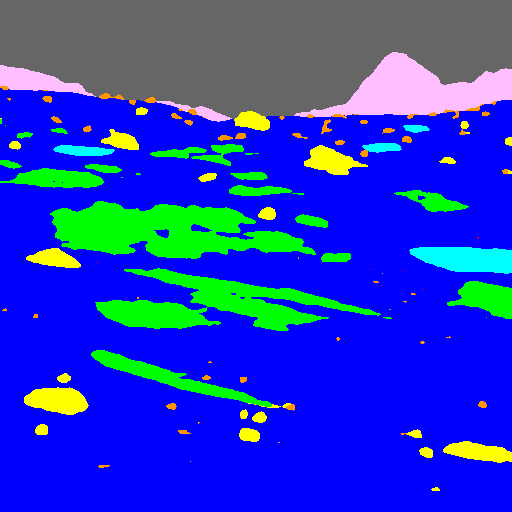}      &
    \includegraphics[width=0.13\linewidth]{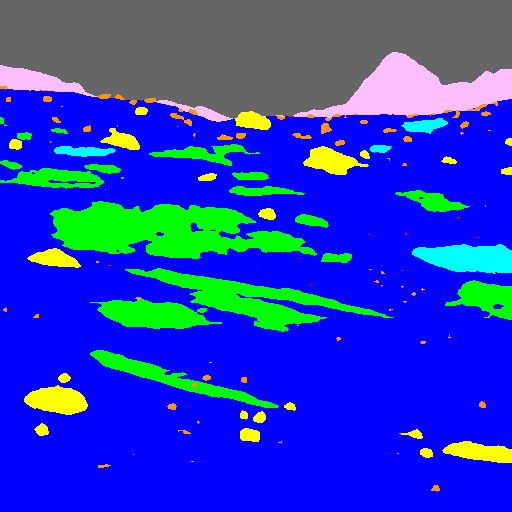}       &
    \includegraphics[width=0.13\linewidth]{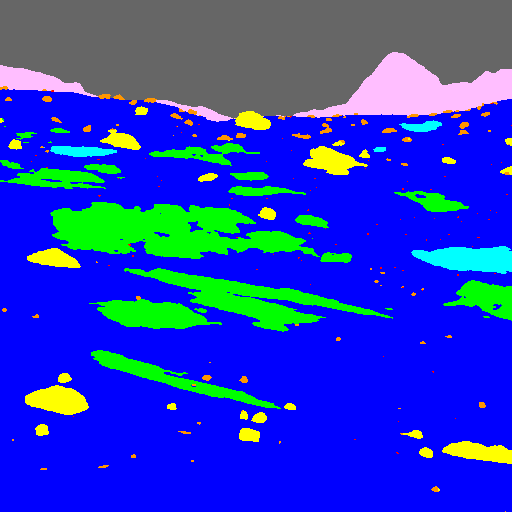}       &
    \includegraphics[width=0.13\linewidth]{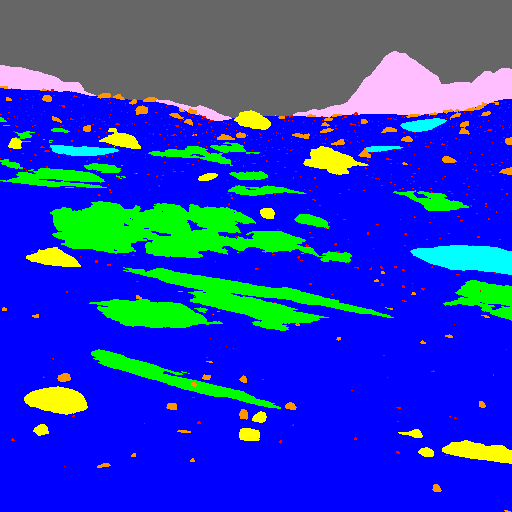} &
    \includegraphics[width=0.13\linewidth]{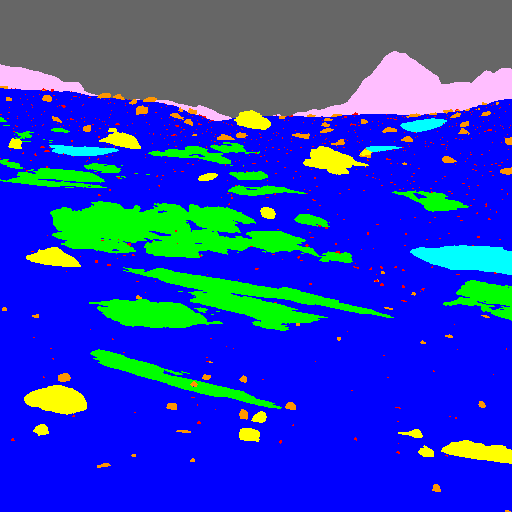}                                                                                   \\
    \includegraphics[width=0.13\linewidth]{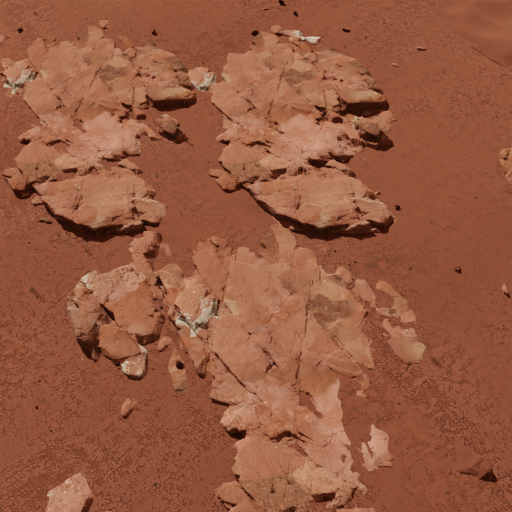}                              &
    \includegraphics[width=0.13\linewidth]{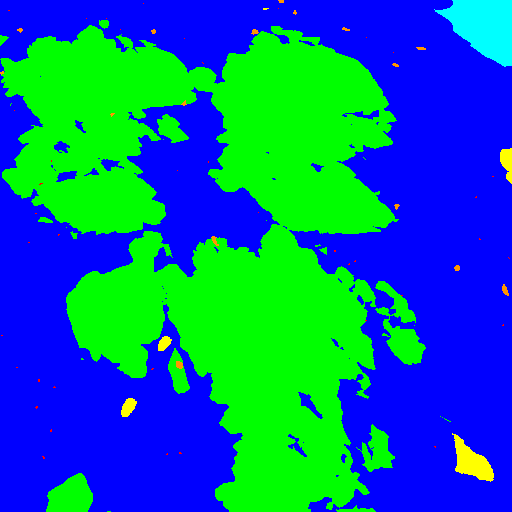}                        &
    \includegraphics[width=0.13\linewidth]{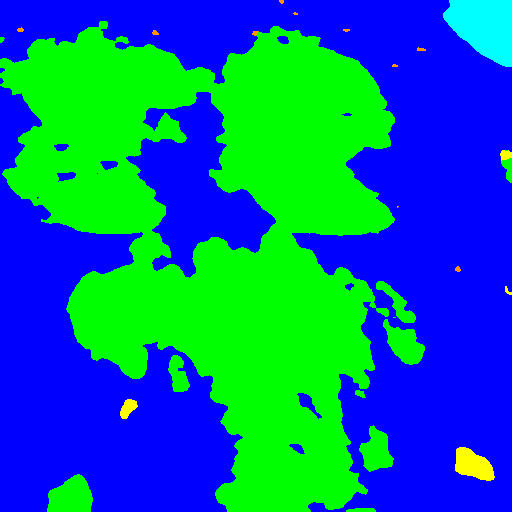}           &
    \includegraphics[width=0.13\linewidth]{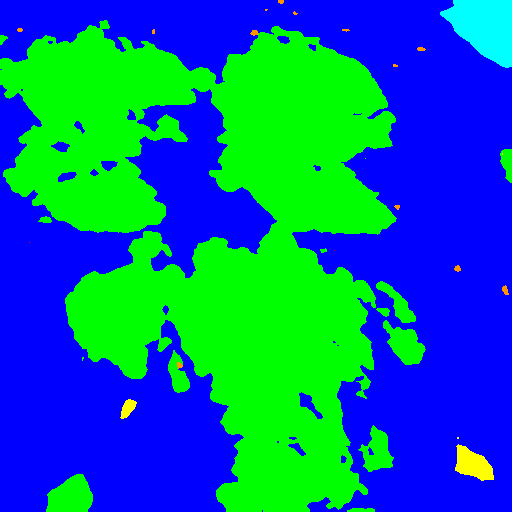}            &
    \includegraphics[width=0.13\linewidth]{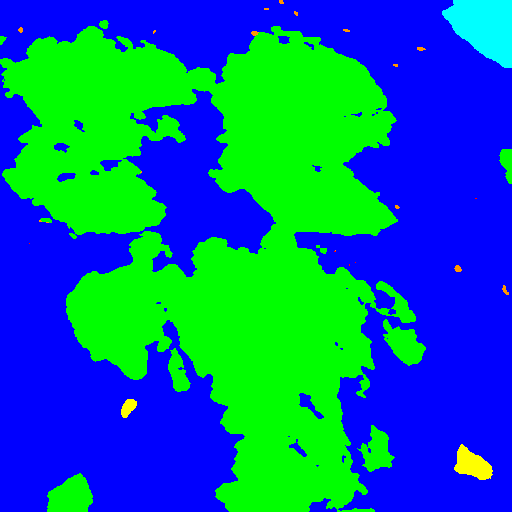}            &
    \includegraphics[width=0.13\linewidth]{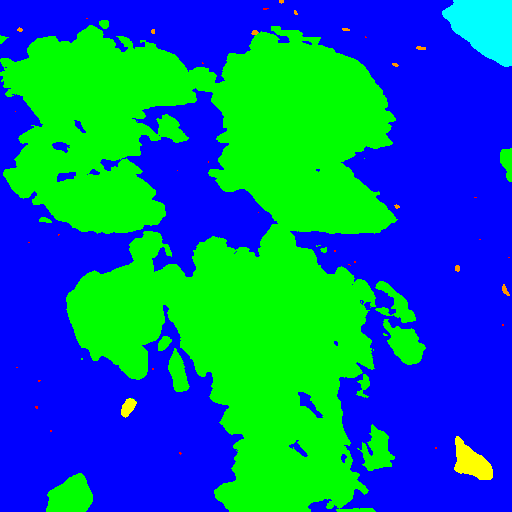}      &
    \includegraphics[width=0.13\linewidth]{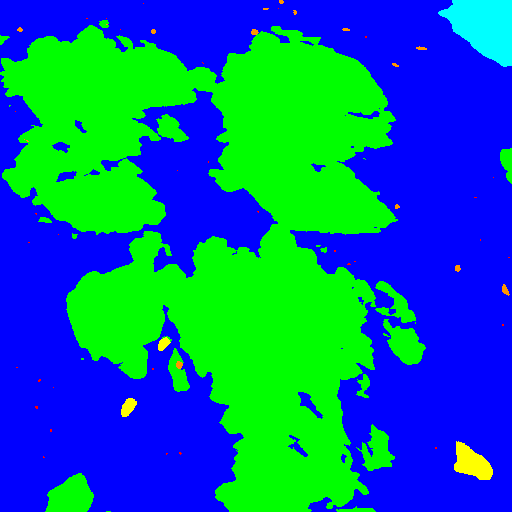}                                                                                        \\
    \includegraphics[width=0.13\linewidth]{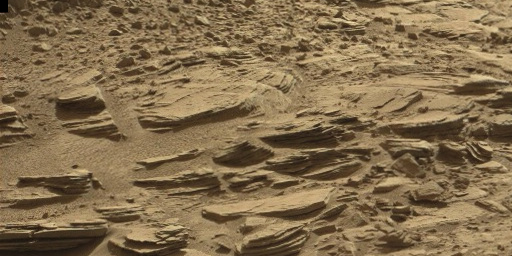}                              &
    \includegraphics[width=0.13\linewidth]{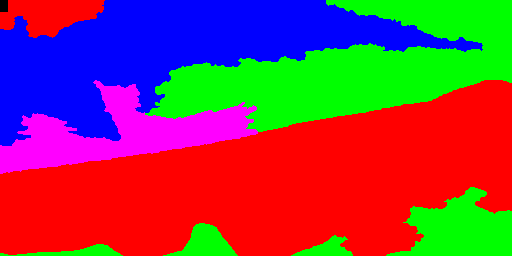}                        &
    \includegraphics[width=0.13\linewidth]{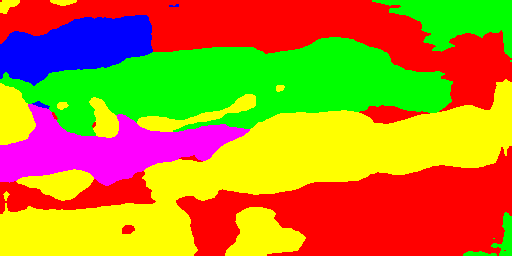}           &
    \includegraphics[width=0.13\linewidth]{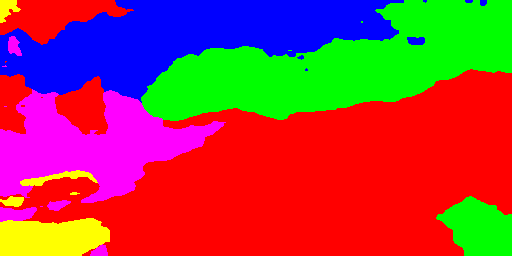}            &
    \includegraphics[width=0.13\linewidth]{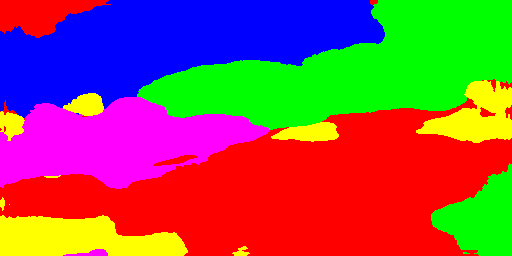}            &
    \includegraphics[width=0.13\linewidth]{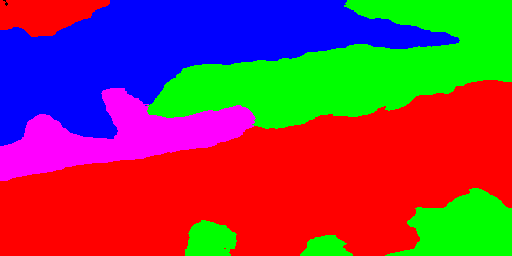}      &
    \includegraphics[width=0.13\linewidth]{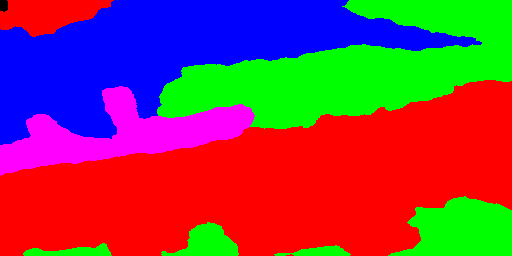}                                                                                        \\
    Image                                                                                                                    & Ground Truth & DeepLabV3+ & SegFormer-B0 & Light4Mars-B & \nnWNet{} & \nnMNet{}
  \end{tabular}

  \caption{Qualitative results on SynMars-TW (top), SynMars-Air (middle), and MarsScapes (bottom).}
  \label{fig:qualtitative_results}
\end{figure*}

We provide a qualitative comparison of our \nnMNet{} against Light4Mars-B and \nnWNet{} on our benchmark.
As illustrated in Fig.~\ref{fig:qualtitative_results}, our model better captures small objects and predicts the correct class and yields accurate predictions on highly unstructured objects.

\subsubsection{Heatmap of CNN and Transformer Branches}
\begin{figure}[t]
  \centering
  \footnotesize
  \setlength{\tabcolsep}{1pt}
  \begin{tabular}{ccccc}
    \includegraphics[width=0.19\linewidth]{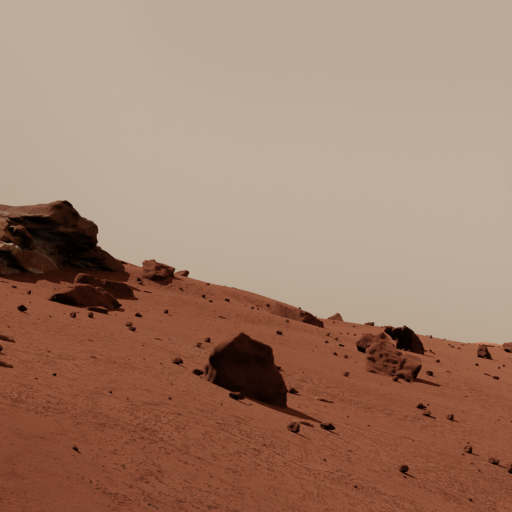}                                 &
    \includegraphics[width=0.19\linewidth]{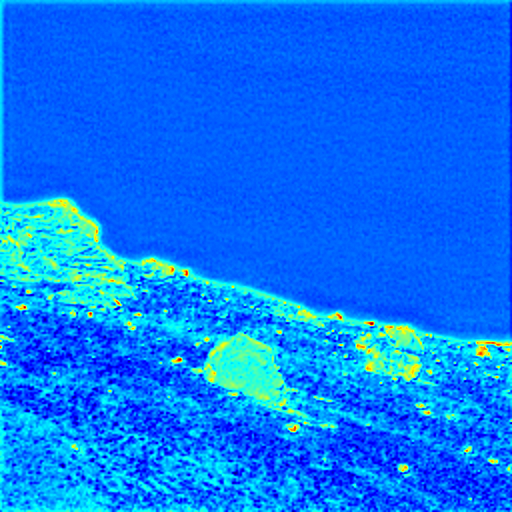}  &
    \includegraphics[width=0.19\linewidth]{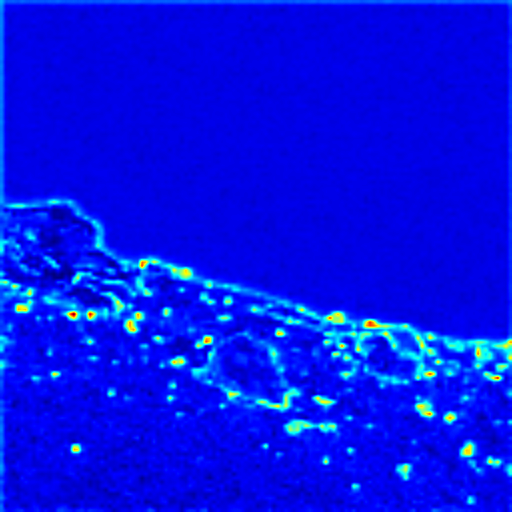}  &
    \includegraphics[width=0.19\linewidth]{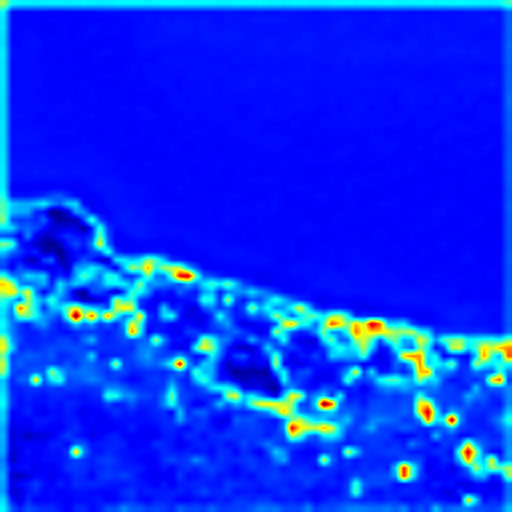}  &
    \includegraphics[width=0.19\linewidth]{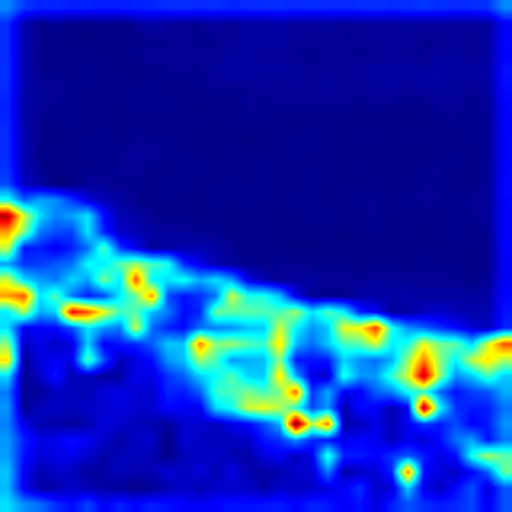}                                          \\

    \includegraphics[width=0.19\linewidth]{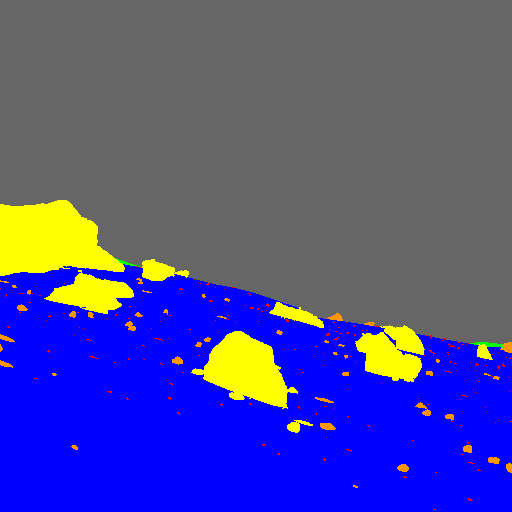}                           &
    \includegraphics[width=0.19\linewidth]{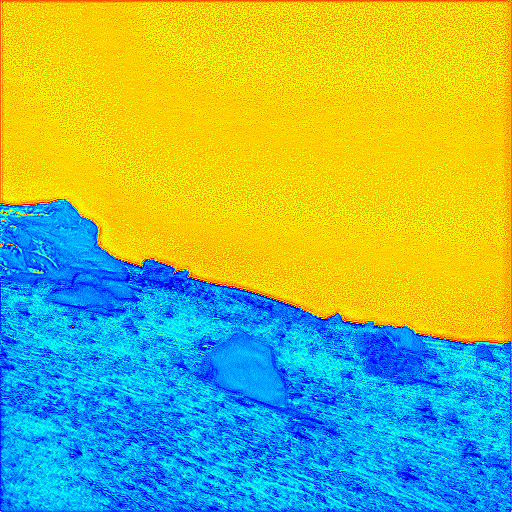} &
    \includegraphics[width=0.19\linewidth]{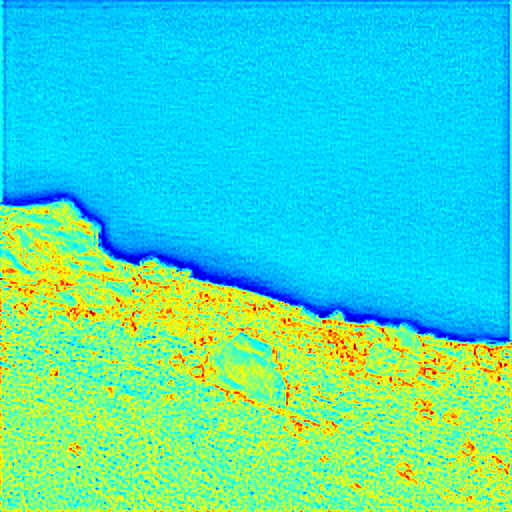} &
    \includegraphics[width=0.19\linewidth]{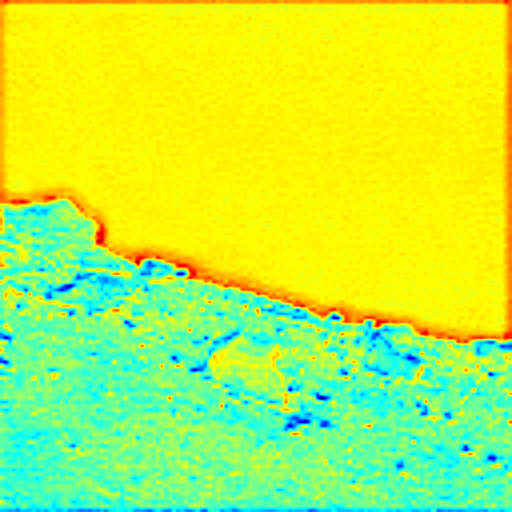} &
    \includegraphics[width=0.19\linewidth]{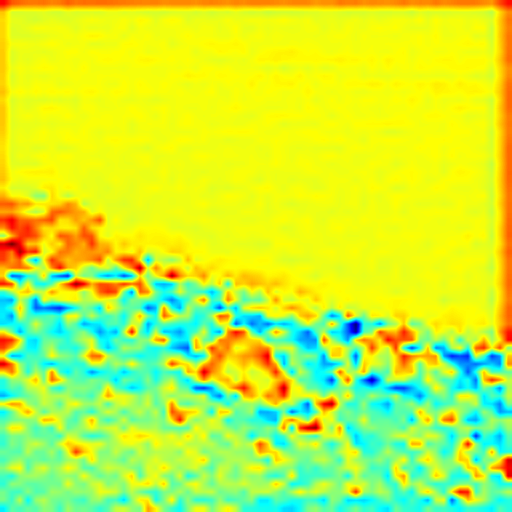}                                         \\
    Image / GT                                                                                                                       & Stage 1 & Stage 2 & Stage 3 & Stage 4
  \end{tabular}
  \caption{
    Heatmap visualization of the encoder (top) and the bridges (down). The former focuses on local details, while the latter captures global context.}
  \label{fig:local_global}
\end{figure}

Fig.~\ref{fig:local_global} illustrates the heatmaps of the convolutional and Transformer blocks, implemented via IRBs and LABs, respectively. IRBs focus on fine-grained local details and exhibit highly concentrated activation maps. In contrast, LABs display broader high-activation regions that span multiple stages. This visualization confirms that our model successfully leverages the distinct advantages of both block types, effectively combining local and global features to enrich the final feature representation.

\subsubsection{Heatmap of Transformer Blocks}
\begin{figure}[t]
  \centering
  \footnotesize
  \setlength{\tabcolsep}{1pt}
  \begin{tabular}{ccccc}
    \includegraphics[width=0.19\linewidth]{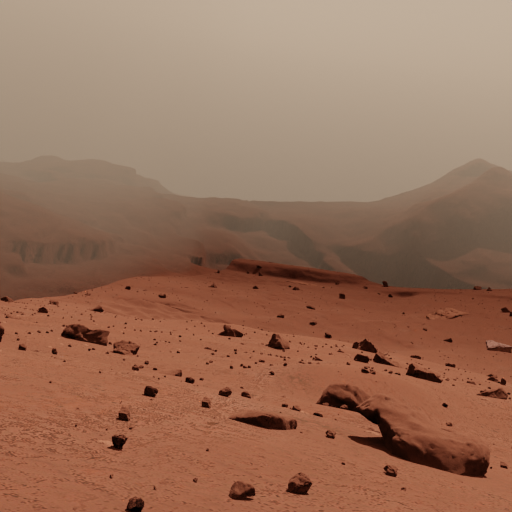}                                 &
    \includegraphics[width=0.19\linewidth]{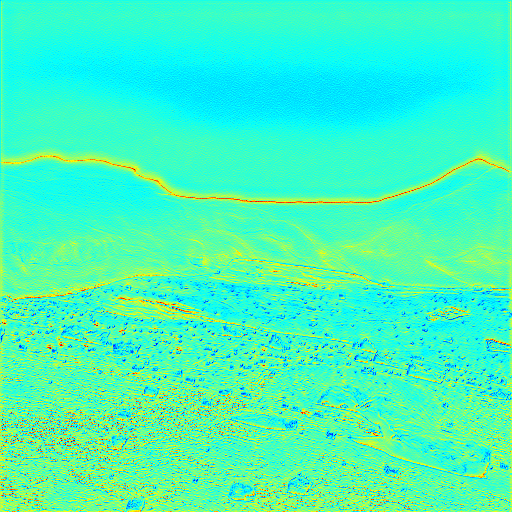} &
    \includegraphics[width=0.19\linewidth]{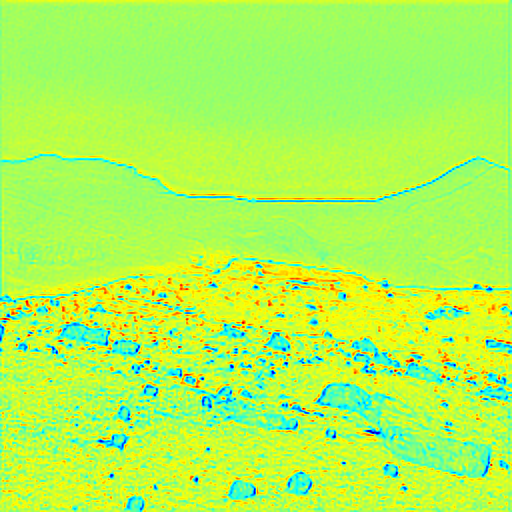} &
    \includegraphics[width=0.19\linewidth]{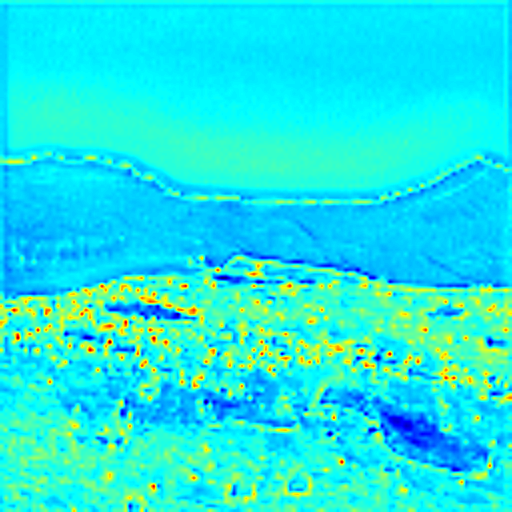} &
    \includegraphics[width=0.19\linewidth]{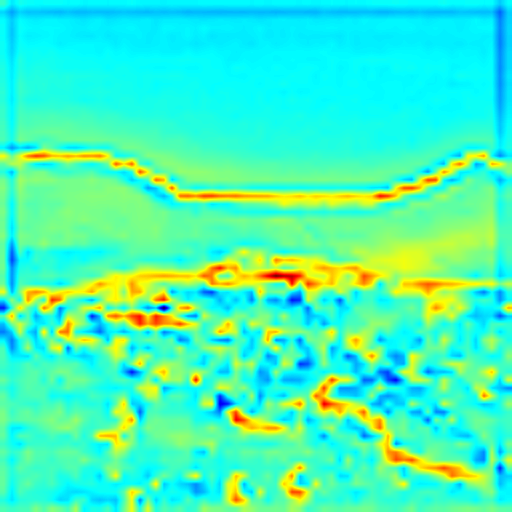}                                         \\

    \includegraphics[width=0.19\linewidth]{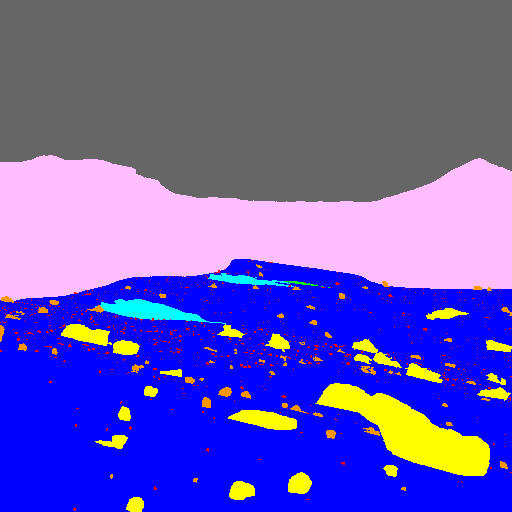}                           &
    \includegraphics[width=0.19\linewidth]{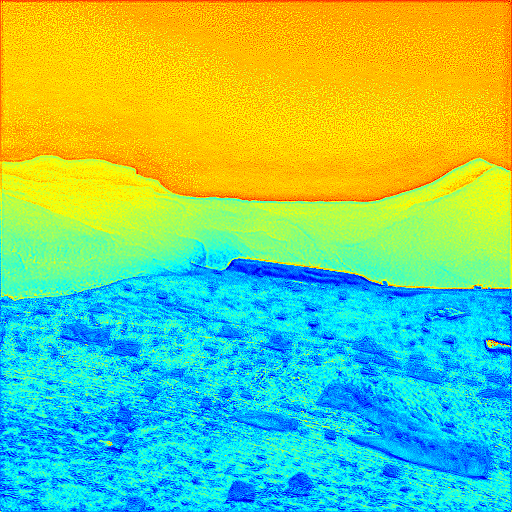} &
    \includegraphics[width=0.19\linewidth]{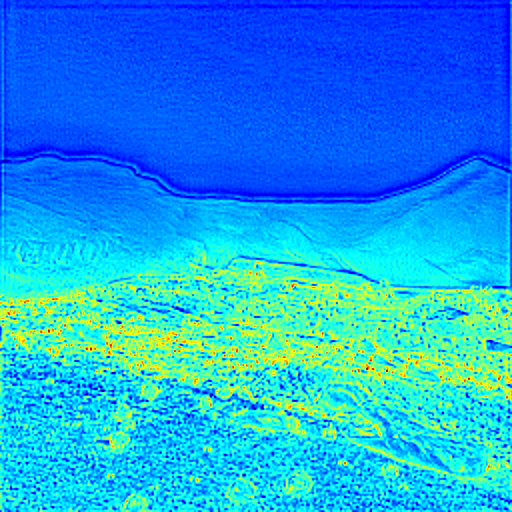} &
    \includegraphics[width=0.19\linewidth]{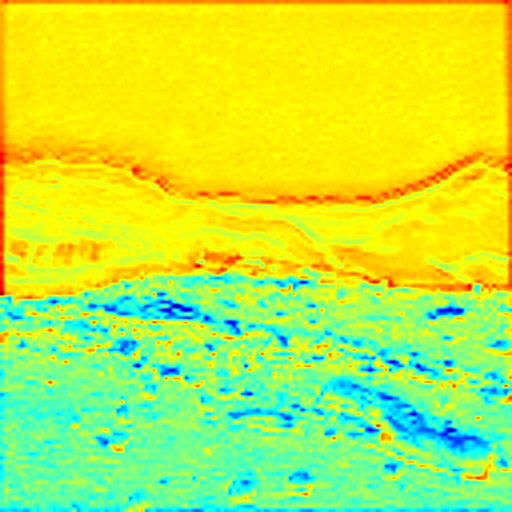} &
    \includegraphics[width=0.19\linewidth]{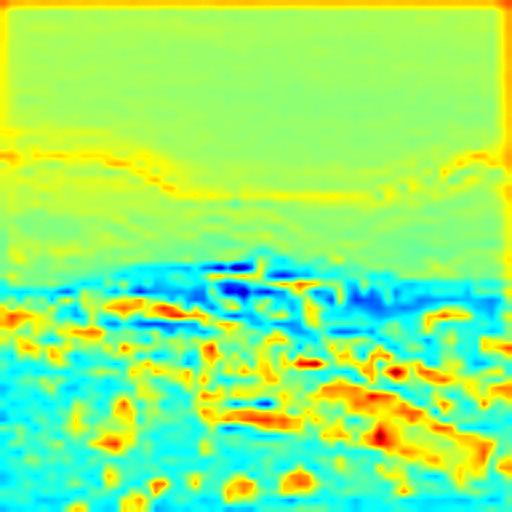}                                         \\
    Image / GT                                                                                                                       & Stage 1 & Stage 2 & Stage 3 & Stage 4
  \end{tabular}
  \caption{
    Heatmap visualization of the bridges in \nnWNet{} (top) and our \nnMNet{} (down). Our method shows greater contrast.}
  \label{fig:global_global}
\end{figure}

Fig.~\ref{fig:global_global} compares the heatmaps of the Transformer blocks in \nnWNet{} and our \nnMNet{}.
By replacing average pooling with RALA, our model demonstrates higher contrast and more pronounced feature representations without losing focus in the deeper layers.
In contrast, \nnWNet{} shows relatively uniform, lower-contrast activations throughout all stages.

\section{Conclusion}
We propose \nnMNet{}, a new baseline model for Martian terrain semantic segmentation that improves upon \nnWNet{} through three architectural enhancements: a Linear Attention Block (LAB) for efficient hybrid CNN-Transformer design, an improved Inverted Residual Block (IRB) for reduced computational cost and enhanced local perception, and a Spatially-Aware Fusion Block (SAFB) for pixel-wise local-global feature fusion via cross-channel modulation. When integrated into the nnUNet framework and evaluated on a unified benchmark of three large-scale datasets, \nnMNet{} achieves new state-of-the-art results, consistently outperforming existing methods. One notable limitation is inference speed. Although \nnMNet{} has fewer parameters and FLOPs than \nnWNet{}, it is slower in terms of inference speed. In contrast, our single-stage variant, \nnMNetPlus{}, achieves higher computational efficiency but still lags behind other methods. Future work will focus on developing hardware-efficient architectures for onboard deployment on Mars rovers.




\bibliographystyle{IEEEtran}
\bibliography{bibliography}







\vfill

\end{document}